\documentclass{article}
\pdfoutput=1

\usepackage[nonatbib,final] {neurips_2019}
\usepackage{amsmath}
\usepackage{bbm}
\usepackage{amsfonts}
\usepackage{mathtools}
\usepackage{booktabs}
\usepackage[hidelinks]{hyperref}
\usepackage[pdftex,dvipsnames]{xcolor}  %
\usepackage{subcaption}
\usepackage{enumitem}%
\usepackage{balance}
\usepackage{multicol}

\usepackage{caption}

\usepackage[colorinlistoftodos,textwidth=20mm,prependcaption,textsize=tiny]{todonotes}

\newcommand{\ourgan}{ScratchGAN}

\newcommand{\tdist}{p^\ast\!} %
\newcommand{\tdistx}{\tdist(\vx)}

\newcommand{\modeldistx}{p_{\vtheta}(\vx)}

\newcommand\cut[1]{}

\newcommand{\squishlist}{
   \begin{list}{$\bullet$}
    { \setlength{\itemsep}{0pt}      \setlength{\parsep}{3pt}
      \setlength{\topsep}{3pt}       \setlength{\partopsep}{0pt}
      \setlength{\leftmargin}{1.5em} \setlength{\labelwidth}{1em}
      \setlength{\labelsep}{0.5em} } }

\newcommand{\squishlisttwo}{
   \begin{list}{$\bullet$}
    { \setlength{\itemsep}{0pt}    \setlength{\parsep}{0pt}
      \setlength{\topsep}{0pt}     \setlength{\partopsep}{0pt}
      \setlength{\leftmargin}{2em} \setlength{\labelwidth}{1.5em}
      \setlength{\labelsep}{0.5em} } }

\newcommand{\squishend}{
    \end{list}  }

\newcommand{\myvec}[1]{\mathbf{#1}}
\newcommand{\myvecsym}[1]{\boldsymbol{#1}}

\newcommand{\vphi}{\myvecsym{\phi}}

\newcommand{\vtheta}{\myvecsym{\theta}}

\newcommand{\vx}{\myvec{x}}

\newcommand{\E}{\mathbb{E}}

\newcommand{\be}{\begin{equation}}
\newcommand{\ee}{\end{equation}}
\newcommand{\bea}{\begin{eqnarray}}
\newcommand{\eea}{\end{eqnarray}}
\newcommand{\beaa}{\begin{eqnarray*}}
\newcommand{\eeaa}{\end{eqnarray*}}

\DeclareMathOperator*{\argmax}{arg\,max}
\DeclareMathAlphabet{\mathpzc}{OT1}{pzc}{m}{n}

\newcommand{\disc}{\mathcal{D}_{\vphi}}

\usepackage[numbers,sort]{natbib}

\graphicspath{{figures/}, }

\usepackage{array}
\newcolumntype{L}[1]{>{\raggedright\let\newline\\\arraybackslash\hspace{0pt}}m{#1}}
\newcolumntype{C}[1]{>{\centering\let\newline\\\arraybackslash\hspace{0pt}}m{#1}}
\newcolumntype{R}[1]{>{\raggedleft\let\newline\\\arraybackslash\hspace{0pt}}m{#1}}

\presetkeys{todonotes}{inline}{}

\title{Training Language GANs from Scratch}

\author{
Cyprien de Masson d'Autume\thanks{Equal contribution.} \quad Mihaela Rosca\footnotemark[1] \quad
Jack Rae \quad Shakir Mohamed\\
  DeepMind\\
  \texttt{\{cyprien,mihaelacr,jwrae,shakir\}@google.com}
}

\begin{document}

\maketitle
\begin{abstract}

Generative Adversarial Networks (GANs) enjoy great success at image generation, but have proven difficult to train in the domain of natural language. Challenges with gradient estimation, optimization instability, and mode collapse have lead practitioners to resort to maximum likelihood pre-training, followed by small amounts of adversarial fine-tuning. The benefits of GAN fine-tuning for language generation are unclear, as the resulting models produce comparable or worse samples than traditional language models. We show it is in fact possible to train a language GAN from scratch --- without maximum likelihood pre-training. We combine existing techniques such as large batch sizes, dense rewards and discriminator regularization to stabilize and improve language GANs. The resulting model,~\ourgan, performs comparably to maximum likelihood training on EMNLP2017 News and WikiText-103 corpora
according to quality and diversity metrics.
\end{abstract}

\section{Introduction}

Unsupervised word level text generation is a stepping stone for a plethora of applications,
from dialogue generation to machine translation and
summarization~\citep{mt,unsupmt,dg,summarization}.
While recent innovations such as architectural changes and leveraging big datasets
are promising~\citep{transformer,jozefowicz2016exploring,radford2019language},
the problem of unsupervised text generation is far from being solved.

Today, language models trained using maximum likelihood are the most successful and widespread approach to text modeling, but they are not without limitations.
Since they explicitly model sequence probabilities,
language models trained by maximum likelihood are
often confined to an autoregressive structure, limiting
applications such as one-shot language generation.
Non-autoregressive maximum likelihood models have been proposed,
but due to reduced model capacity they
rely on distilling autoregressive models to achieve comparable performance
on machine translation tasks
~\citep{gu2017non}.

When combined with maximum likelihood training, autoregressive modelling
can result in poor samples due exposure bias~\citep{bengio2015scheduled}--
a distributional shift between
training sequences used for learning and model data required for generation.
Recently,~\citep{nucleus_sampling} showed that sampling from state of the art language models can lead to
repetitive, degenerate output.
Scheduled sampling~\citep{bengio2015scheduled} has been proposed as
a solution, but is thought to encourage sample quality by reducing sample diversity, inducing mode collapse~\citep{huszar2015not}.

Generative Adversarial Networks (GANs)~\citep{gan} are an alternative to models trained via maximum likelihood.
GANs do not suffer from exposure bias
since the model learns to sample during training: the learning objective is to generate samples which
are indistinguishable from real data according to a discriminator.
Since GANs don't require an explicit probability model, they remove the restriction to autoregressive architectures, allowing one shot feed-forward generation~\citep{improvedwgan}.

The sequential and discrete nature of text has made the application of GANs to language challenging, with fundamental issues such as difficult gradient estimation and mode collapse yet to be addressed. Existing language GANs avoid these issues by pre-training models with maximum likelihood
~\citep{seqgan,leakygan,textgan,maligan,rankgan}
and limiting the amount of adversarial fine tuning
by restricting the number of fine-tuning epochs and often using
a small learning rate \citep{semeniuta2018accurate, fallingshort}.
This suggests ``that the best-performing GANs tend to
stay close to the solution given by maximum-likelihood
training"~\citep{fallingshort}.
Even with adversarial fine-tuning playing a limited role, extensive evaluation
has shown that existing language GANs do not improve over maximum likelihood-trained
models~\citep{semeniuta2018accurate,fallingshort,tevet2018evaluating}.

We show that pure adversarial training is a viable approach for unsupervised word-level text generation by training a language GAN from scratch. We achieve this by tackling the fundamental limitations of training discrete GANs through a combination of existing techniques as well as carefully choosing the model and training regime.
To the best of our knowledge we are the first to do so successfully; we thus call our model~\ourgan.
Compared to prior work on discrete language GANs which ``barely achieve non-random results without supervised pre-training"~\citep{semeniuta2018accurate},
~\ourgan~achieves results comparable with maximum likelihood models.

Our aim is to learn models that captures both both semantic coherence and grammatical
correctness of language, and to demonstrate that these properties have been captured
with the use of different evaluation metrics. BLEU and
Self-BLEU~\citep{zhu2018texygen} capture basic local consistency. The Fr\'echet Distance
metric~\citep{semeniuta2018accurate} captures global consistency and semantic information, while
being less sensitive to local syntax. We use Language and Reverse Language model
scores~\citep{fallingshort} across various softmax temperatures to capture the diversity-quality
trade-off. We measure validation data perplexity, using the fact that ~\ourgan~ learns
an explicit distribution over sentences.
Nearest neighbor analysis in embedding and data space provide evidence that our model is not trivially overfitting, e.g. by copying sections of training text.

We make the following contributions:
\vspace{-2mm}
\begin{itemize}[leftmargin=*,noitemsep]

\item We show that GANs without any pre-training are comparable with maximum likelihood methods at unconditional text generation.
\item We show that large batch sizes, dense rewards and discriminator regularization are key ingredients of training language GANs from scratch.

\item We perform an extensive evaluation of the quality and diversity of our model. In doing so, we show that no current evaluation metric is able to capture all the desired properties of language.
\end{itemize}

The ScratchGAN code can be found at \url{https://github.com/deepmind/deepmind-research/scratchgan}.

\section{Generative Models of Text}

The generative model practitioner has two choices to make: how to model the unknown data distribution $\tdistx$ and how to learn the parameters $\vtheta$ of the model. The choice of model is where often prior information about the data is encoded, either through the factorization of the distribution, or through its parametrization.
The language sequence $\vx = [x_1, ..., x_{T}]$ naturally lends itself to autoregressive modeling:
\vspace{-1mm}
\begin{equation}
  p_{\vtheta}(\vx) = \prod_{t=1}^T p_{\vtheta}(x_t |x_1, ..., x_{t-1})
\label{eq:autoregressive}
\end{equation}
Sampling $\hat{x}_1, ...,\hat{x}_T$ from an autoregressive model is an iterative process:
each token $\hat{x}_t$ is sampled from the conditional distribution imposed by previous samples: $\hat{x}_t \sim p_{\vtheta}(x_t |\hat{x}_1, ..., \hat{x}_{t-1})$.
Distributions $p_{\vtheta}(x_t |x_1, ..., x_{t-1})$ are Categorical distributions over the vocabulary size, and are often parametrized as recurrent neural networks~\citep{lstm,gru}.

The specific tokenization $x_1, ..., x_{T}$ for a given data sequence is left to the practitioner, with character level or word level splits being the most common.
Throughout this work, we use word level language modeling.

\subsection{Maximum Likelihood}
Once a choice of model is made, the question of how to \textit{train} the model arises. The most
common approach to learn model of language is using maximum likelihood estimation (MLE):
\begin{equation}
  \argmax_{\vtheta} \mathbb{E}_{\tdistx} \log p_\theta(\vx)
\label{eq:mle}
\end{equation}
The combination of autoregressive models and maximum likelihood learning has been very fruitful in language modeling~\citep{shannon1951prediction,mikolov2010recurrent,transformer}, but it is unclear whether maximum likelihood is the optimal perceptual objective for text data~\citep{huszar2015not}.
In this work we will retain the use of autoregressive models and focus on the impact of the training
criterion on the quality and sample diversity of generated data, by using adversarial training instead.

\subsection{Generative Adversarial Networks}
Generative adversarial networks~\citep{gan} learn the data distribution $\tdistx$ through a two player adversarial game between a discriminator and a generator.
A discriminator $\mathcal{D}_\phi(\vx)$ is trained to distinguish between real data and
samples from the generator distribution $p_{\vtheta}(\vx)$, while the generator is trained to fool the discriminator in identifying its samples as real.
The original formulation proposes a min-max optimization procedure using the objective:
\begin{equation}
 \min_{\vtheta} \max_{\vphi} \mathbb{E}_{\tdistx}\bigl[\log \disc(\vx)\bigr] +
 \E_{p_{\vtheta}(\vx)}\bigl[\log (1 -\disc(\vx))\bigr].
 \label{eq:gan_min_max}
\end{equation}
\citet{gan} suggested using the \textit{alternative generator loss} $\E_{\modeldistx}[-\log \disc(x)]$ as it provides better gradients for the generator. Since then, multiple other losses have been proposed \citep{wgan, lsgan, relativisticgan,implicitlearning}.

Challenges of learning language GANs arise from the combination of the adversarial learning principle
with the choice of an autoregressive model.
Learning $p_{\vtheta}(\vx) = \prod_{t=1}^T p_{\vtheta}(x_t |x_1, ..., x_{t-1})$ using equation~\ref{eq:gan_min_max} requires backpropagating through a sampling operation, forcing the language GAN practitioner to choose between high variance, unbiased estimators such as REINFORCE~\citep{reinforce}, or lower variance, but biased estimators, such as the Gumbel-Softmax trick~\citep{gumbel_softmax_concrete, gumbel_softmax} and other continuous relaxations~\citep{improvedwgan}.
Gradient estimation issues compounded with other GAN problems such as mode collapse or training instability~\citep{wgan, manypaths} led prior work on language GANs to use maximum likelihood pre-training~\citep{seqgan,leakygan,maskgan,relgan,textgan,rankgan}. This is the current preferred approach to train text GANs.

\subsection{Learning Signals}

To train the generator we use the REINFORCE gradient estimator~\citep{reinforce}:
\begin{equation}
  \nabla_{\theta} \mathbb{E}_{p_\theta(\vx)}[R(\vx)]
 = \mathbb{E}_{p_\theta(\vx)} \bigl[R(\vx) \nabla_{\theta} \log p_\theta(\vx)\bigr],
\label{eq:pg}
\end{equation}
where $R(\vx)$ is provided by the discriminator.
By analogy with reinforcement learning, we call $R(\vx)$ a \textit{reward}.
Setting $R(\vx) = \frac{p^*(\vx)}{p_\theta(\vx)}$, recovers the MLE estimator in Eq~\eqref{eq:mle} as
shown by \citet{maligan}:
\vspace{-1mm}
\begin{equation}
\mathbb{E}_{p_\theta(\vx)}\left[ \frac{p^*(\vx)}{p_\theta(\vx)} \nabla_{\theta} \log p_\theta(\vx) \right] =
  \mathbb{E}_{p*(\vx)} \bigl[ \nabla_{\theta} \log p_\theta(\vx) \bigr] = \nabla_{\theta}
  \mathbb{E}_{p*(\vx)} \log p_\theta(\vx).
\end{equation}
The gradient updates provided by  the MLE estimator can be seen as a special case of
the REINFORCE updates used in language GAN training.
The important difference lies in the fact that for language GANs rewards are learned.
Learned discriminators have been shown to be a useful measure of model quality and correlate with human evaluation~\citep{orioladvdialogue}.
We postulate that learned rewards provide a smoother signal to the generator than the classical MLE loss: the discriminator can learn to generalize and provide a meaningful signal over parts of the distribution not covered by the training data.
As the training progresses and the signal from the discriminator improves, the generator also explores other parts of data space, providing a natural curriculum, whereas MLE models are only exposed to the dataset.

Adversarial training also enables the use of domain knowledge. Discriminator ensembles where each
discriminator is biased to focus on specific aspects of the samples such as syntax, grammar,
semantics, or local versus global structure are a promising approach~\citep{holtzman2018learning}.
The research avenues opened by learned rewards and the issues with MLE pre-training motivate our search for a language GAN which does not make use of maximum likelihood pre-training.

\section{Training Language GANs from Scratch}
\label{sec:method}

To achieve the goal of training a language GAN from scratch, we tried
different loss functions and architectures, various reward structures and regularization methods, ensembles, and other modifications.
Most of these approaches did not succeed or did not result in any significant gains.
Via this extensive experimentation we found that the key ingredients to train language GANs from scratch are:
a recurrent discriminator used to provide dense rewards at each time step, large batches for variance reduction, and discriminator regularization.
We describe the generator architecture and reward structure we found effective in Figure~\ref{fig:model} and
provide a list of other techniques we tried but which proved unsuccessful or unnecessary in Appendix~\ref{app:neg_res}.

\begin{figure*}[t]
\begin{minipage}[t]{\columnwidth}
\begin{minipage}{0.57\columnwidth}
		\centering
		\includegraphics[width=\columnwidth]{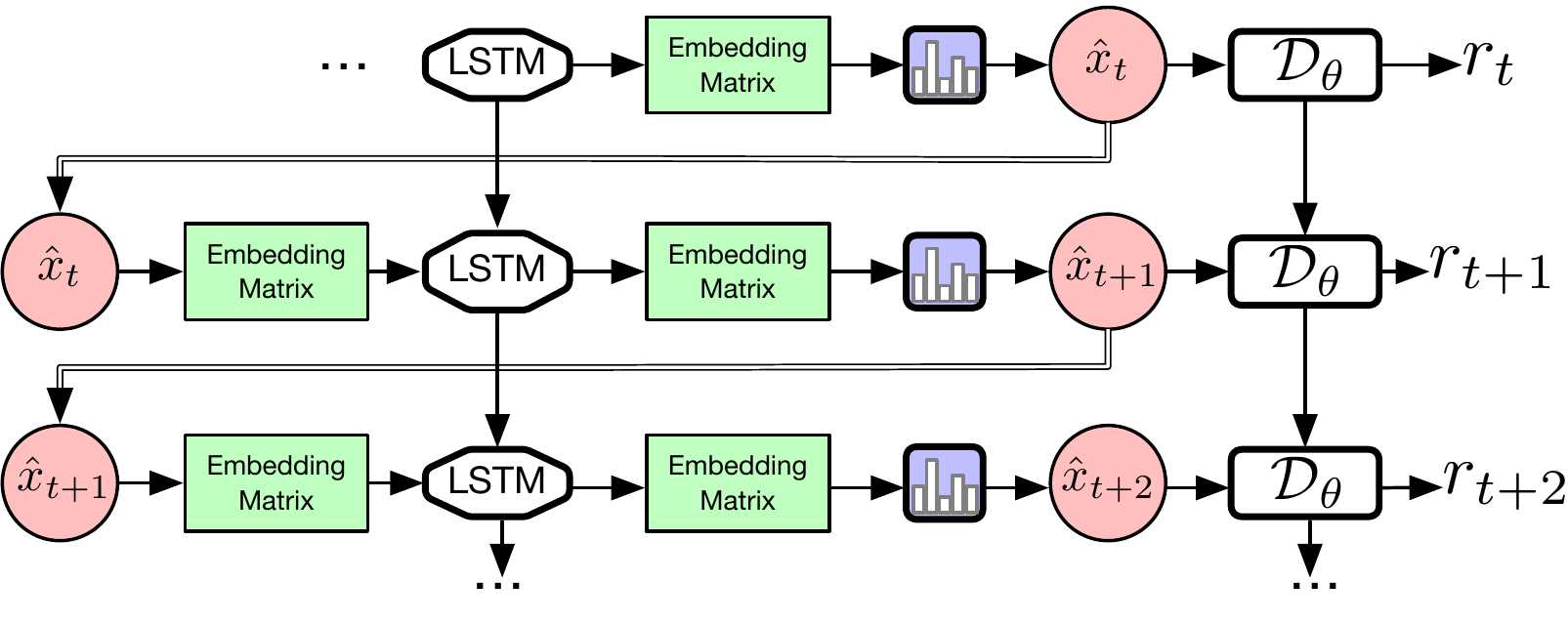}
	\captionof{figure}{~\ourgan~architecture and reward structure.}
\label{fig:model}
\end{minipage}
\hspace{5mm}
\begin{minipage}{0.4\columnwidth}

	\vspace{10mm}
	\begin{center}
			\captionof{table}{BLEU-5 and Self-BLEU-5 metrics for a 5-gram model.}
	\label{tab:bleu5kn}
		\begin{small}
			\begin{sc}
				\begin{tabular}{lcc}
					\toprule
					Model & BLEU-5 & SBLEU-5 \\
					\midrule
					Kneser-Ney & 20.67 & 19.73 \\
					Training data & 20.73 & 20.73 \\
					\bottomrule
				\end{tabular}
			\end{sc}
		\end{small}
	\end{center}
\vspace{3mm}
\end{minipage}
\end{minipage}
\vspace{-4mm}
\end{figure*}

\subsection{Dense Rewards}
Our ultimate goal is to generate entire sequences, so we could
train a discriminator to distinguish between complete data sequences and  complete sampled
sequences, with the generator receiving a reward only after generating a full sequence. However, in
this setting the generator would get no learning signal early in training, when generated sentences can
easily be determined to be fake by the discriminator.
We avoid this issue by instead training a recurrent discriminator which provides rewards for each generated token \citep{maskgan}.
The
discriminator $\disc$ learns to distinguish between sentence prefixes coming from real data and sampled sentence prefixes:
\begin{equation*}
   \max_{\vphi} \sum_{t=1}^{T}\mathbb{E}_{\tdist(x_t |x_1, ..., x_{t-1})}\bigl[\log \disc(x_t | x_1, ...x_{t-1})\bigr]
  + \sum_{t=1}^{T}\mathbb{E}_{p_{\vtheta}(x_t |x_1, ..., x_{t-1})}\bigl[\log (1 - \disc(x_t | x_1,...x_{t-1}))\bigr]
\end{equation*}

While a sequential discriminator is potentially harder to learn than sentence based feed-forward discriminators, it is computationally cheaper than approaches that use Monte Carlo Tree Search to score partial sentences~\citep{leakygan, seqgan, rankgan} and has been shown to perform better empirically~\citep{semeniuta2018accurate}.

For a generated token $\hat{x}_t \sim p_{\vtheta}(x_t| x_{t-1} ... x_1)$, the reward provided to the~\ourgan~generator at time step $t$ is:
\begin{equation}
r_t = 2 \disc(\hat{x}_t| x_{t-1} ... x_1) - 1
\end{equation}
Rewards scale linearly with the probability the discriminator assigns to the current prefix pertaining to a real sentence.
Bounded rewards help stabilize training.

The goal of the generator at timestep $t$ is to maximize the sum of discounted future rewards using a
discount factor $\gamma$:
\vspace{-2mm}
\begin{equation}
R_t = \sum_{s=t}^{T} \gamma^{s-t} r_s
\end{equation}
Like~\ourgan, SeqGAN-step \citep{semeniuta2018accurate} uses a recurrent discriminator to provide rewards per time step to a generator trained using policy gradient for unsupervised word level text generation.
Unlike SeqGAN-step, our model
is trained from scratch using only the adversarial objective, without any maximum likelihood pretraining.

\subsection{Large Batch Sizes for Variance Reduction}

The~\ourgan~generator parameters $\vtheta$ are updated using Monte Carlo estimates of policy gradients (Equation~\ref{eq:pg}), where $N$ is the batch size:
\vspace{-1.5mm}
\begin{gather}
\label{eq:gen_update}
  \nabla_{\theta} = \sum_{n=1}^{N}\sum_{t=1}^{T} (R_t^n - b_t) \nabla_{\theta} \log p_{\vtheta}(\hat{x}_t^n| \hat{x}_{t-1}^n ... \hat{x}_1^n), \qquad
 \hat{x}_t^n \sim p_{\vtheta}(x_t^n| \hat{x}_{t-1}^n ... \hat{x}_1^n)  \nonumber
\end{gather}
A key component of~\ourgan~is the use of large batch sizes to reduce the variance of the gradient estimation,
exploiting the ability to cheaply generate experience by sampling from the generator.
To further reduce the gradient variance~\ourgan~uses a global moving-average of rewards as a baseline $b_t$~\citep{rlbook}, as we empirically found it improves performance for certain datasets.

Providing rewards only for the sampled token as in Equation~\eqref{eq:gen_update} results in a substantial training speed boost compared to methods that use $p_{\vtheta}(x_t^n| \hat{x}_{t-1}^n ... \hat{x}_1^n)$ to provide rewards for each token in the vocabulary, in order to reduce variance and provide a richer learning signal. These methods score all prefixes at time $t$ and thus scale linearly with vocabulary size~\citep{maskgan}.

\subsection{Architectures and Discriminator Regularization}

The~\ourgan~discriminator and generator use an embedding layer followed by one or more LSTM layers~\citep{lstm}. For the embedding layer, we have experimented with training the embeddings from scratch, as well as using pre-trained GloVe embeddings~\citep{glove} concatenated with learned embeddings. When GloVe embeddings are used, they are shared by the discriminator and the generator, and kept fixed during training.

Discriminator regularization in the form of
layer normalization~\citep{layernorm}, dropout~\citep{dropout} and $L_{2}$ weight decay
provide a substantial performance boost to~\ourgan.
Our findings align with prior work which showed the importance of discriminator regularization on image GANs~\citep{biggan, spectralgan, manypaths}.

Despite using a recurrent discriminator, we also provide the discriminator with positional information by concatenating a fix sinusoidal signal to the word embeddings used in the discriminator~\citep{transformer}.
We found this necessary to ensure the sentence length distribution obtained from generator samples matches that of the training data. Ablation experiments are provided in Appendix~\ref{app:pos_info}.

\section{Evaluation Metrics}

\begin{figure}[tb]
\centering
\begin{subfigure}[t]{.45\textwidth}
  \includegraphics[height=4cm]{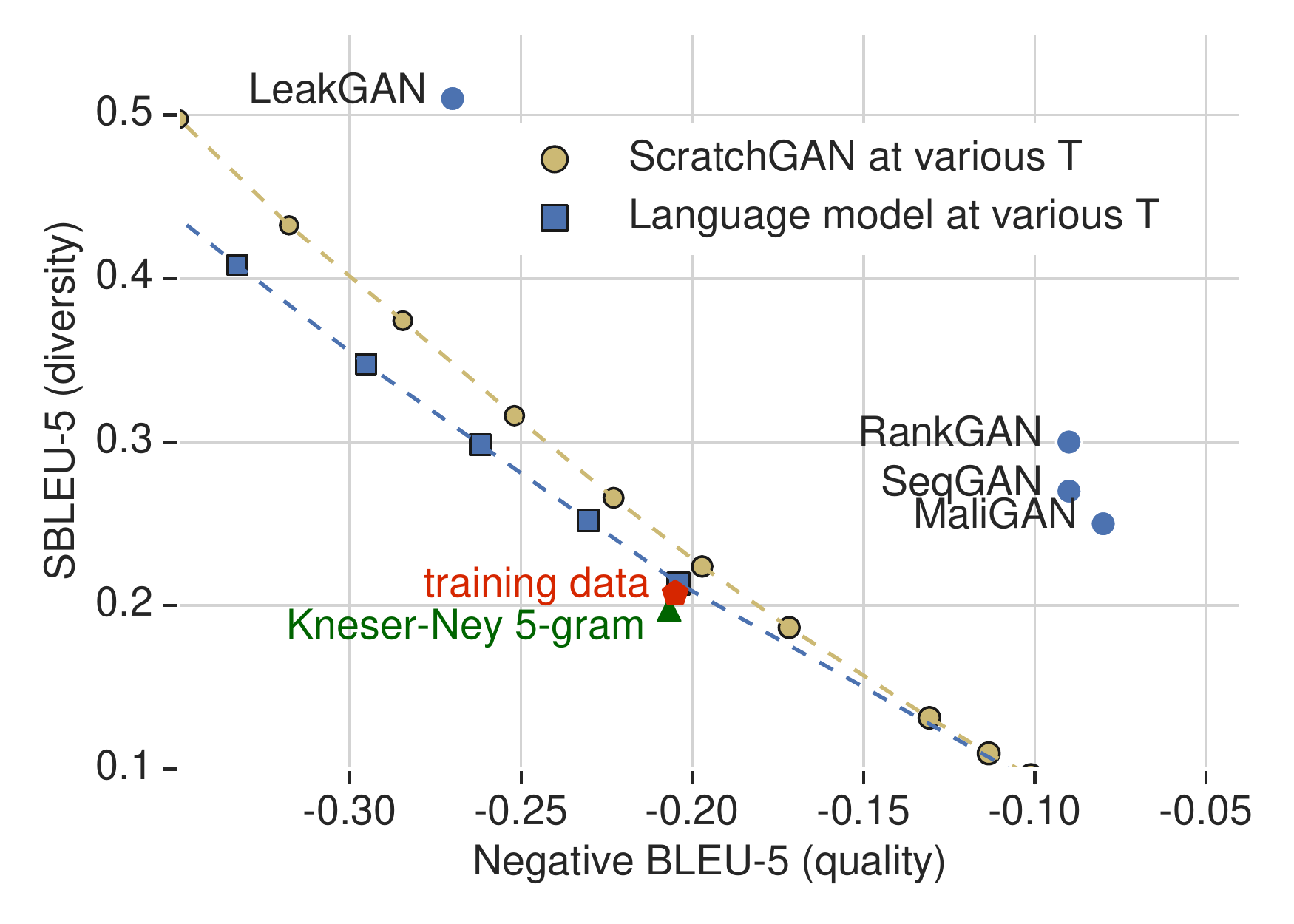}
  \caption{Negative BLEU-5 versus Self-BLEU-5.}
    \label{fig:bleu_metrics}
\end{subfigure}%
\hspace{3mm}
\begin{subfigure}[t]{.5\textwidth}
  \centering
\includegraphics[height=4cm]{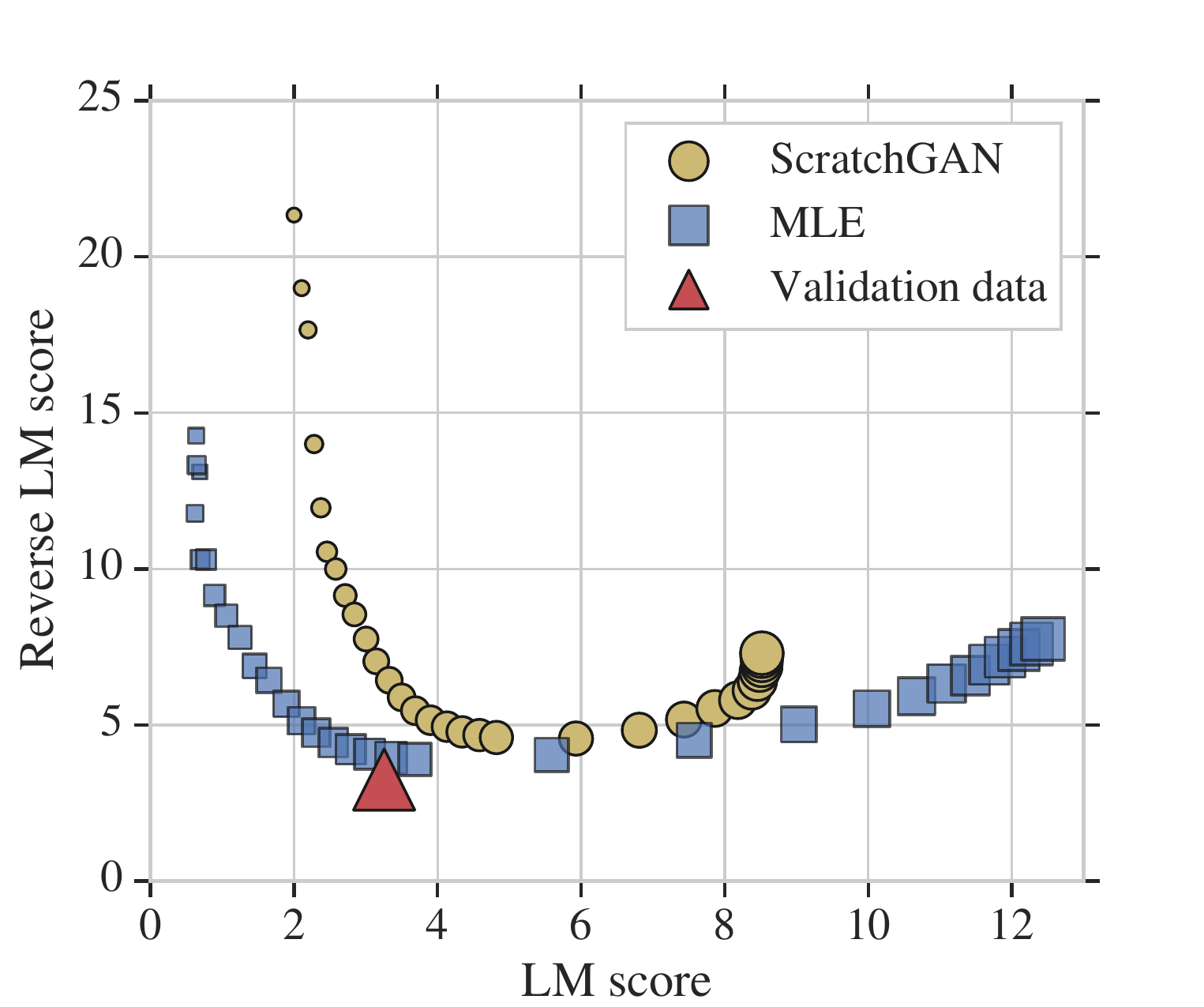}
\caption{Language- and reverse language-model scores.}
\label{fig:wiki_lm_rlm}
\end{subfigure}
\caption{BLEU scores on EMNLP2017 News (left) and language model scores on Wikitext-103 (right).
For BLEU
	scores, left is better and down is better. LeakGAN, MaliGAN, RankGAN and SeqGAN results from
	\citet{fallingshort}.}
\vspace{-4mm}
\end{figure}

Evaluating text generation remains challenging, since no single metric is able to capture all desired
properties:
local and global consistency,
diversity and quality, as well as generalization beyond the training set.
We follow \citet{semeniuta2018accurate} and \citet{fallingshort} in the choice of metrics.
We use $n$-gram based metrics to capture local consistency,
Fr\'echet Distance to measure distances to real data in embedding space, and language model scores to measure the quality-diversity trade-off.
To show our model is not trivially overfitting we look at nearest neighbors in data and embedding space.

\subsection{$n$-gram based Metrics}
BLEU~\citep{papineni2002bleu} and Self-BLEU have been proposed~\citep{zhu2018texygen} as measures of quality and diversity, respectively.
BLEU based metrics capture local consistency and detect relatively simple problems with syntax but do not capture semantic variation~\citep{semeniuta2018accurate,bleuissues}.

We highlight the limitations of BLEU metrics by training a $5$-gram model with
Kneser-Ney smoothing~\citep{kneser1995improved} on EMNLP2017-News and measuring its BLEU
score.
The results are reported in Table~\ref{tab:bleu5kn}.
The $5$-gram model scores close to perfect according to BLEU-5 metric although its samples are qualitatively very poor (see Table~\ref{tab:ngram_samples} in the Appendix).
In the rest of the paper we report BLEU-5 and Self-BLEU-5 metrics to compare with prior work, and
complement it with metrics that capture global consistency, like Fr\'echet Distance.

\subsection{Fr\'echet Embedding Distance}

\citet{semeniuta2018accurate} proposed the Fr\'echet InferSent Distance (FID), inspired by the Fr\'echet Inception Distance used for images~\citep{fid}. The metric computes the Fr\'echet distance between two
Gaussian distributions fitted to data embeddings, and model sample embeddings, respectively. \citet{semeniuta2018accurate} showed that the metric is not sensitive to the choice of embedding model and use InferSent for model evaluation, while we use a Universal Sentence
Encoder~\citep{cer2018universal}\footnote{The model can be found at
\url{https://tfhub.dev/google/universal-sentence-encoder/2}}.
We call the metric Fr\'echet Embedding Distance to clarify that we use a different embedding model from ~\citet{semeniuta2018accurate}.

The Fr\'echet Embedding Distance (FED) offers several advantages over BLEU-based metrics, as highlighted in~
\citet{semeniuta2018accurate}:  it captures both quality and diversity;
it captures global consistency; it is faster and simpler to compute than BLEU metrics; it correlates with human evaluation; it is less sensitive to word order than BLEU metrics; it is empirically proven useful for images.

We find that the Fr\'echet Embedding Distance provides a useful metric to optimize for during model development, and we use it to choose the best models.
However, we notice that FED also has drawbacks: it can be sensitive to sentence length, and we avoid this bias by ensuring that all compared models match the sentence length distribution of the data (see details in Appendix~\ref{app:fd_sen_length}).

\subsection{Language Model Scores}
\citet{fallingshort} proposed evaluating the quality of generated model samples using a language model (Language Model score, LM), as well as training a language model on the generated samples and scoring the original data with it (Reverse Language Model score, RLM).
LM measures sample quality: bad samples score poorly under a language model trained on real data.
RLM measures sample diversity: real data scores poorly under a language model trained on samples which lack diversity.
While insightful, this evaluation criteria relies on training new models, and hence
the results can depend on the evaluator architecture.
The metric could also have an inherent bias favoring language models, since they
were trained using the same criteria.

\begin{figure}[tb]
\centering
\begin{subfigure}{.23\textwidth}
  \centering
  \captionsetup{justification=centering,margin=0cm}
  \includegraphics[width=\linewidth]{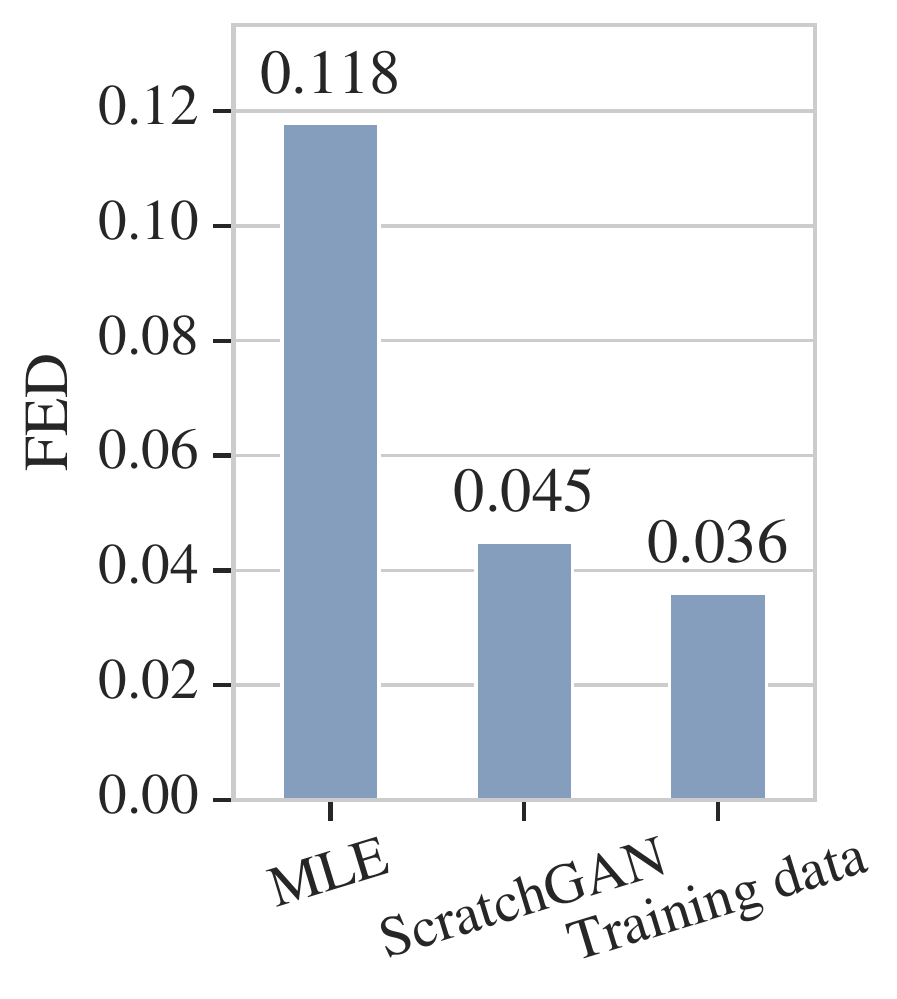}
  \caption{Wikitext-103. }\label{fig:wiki_fd}
\end{subfigure}%
\begin{subfigure}{.3\textwidth}
  \vspace{6mm}
  \centering
  \includegraphics[width=\linewidth]{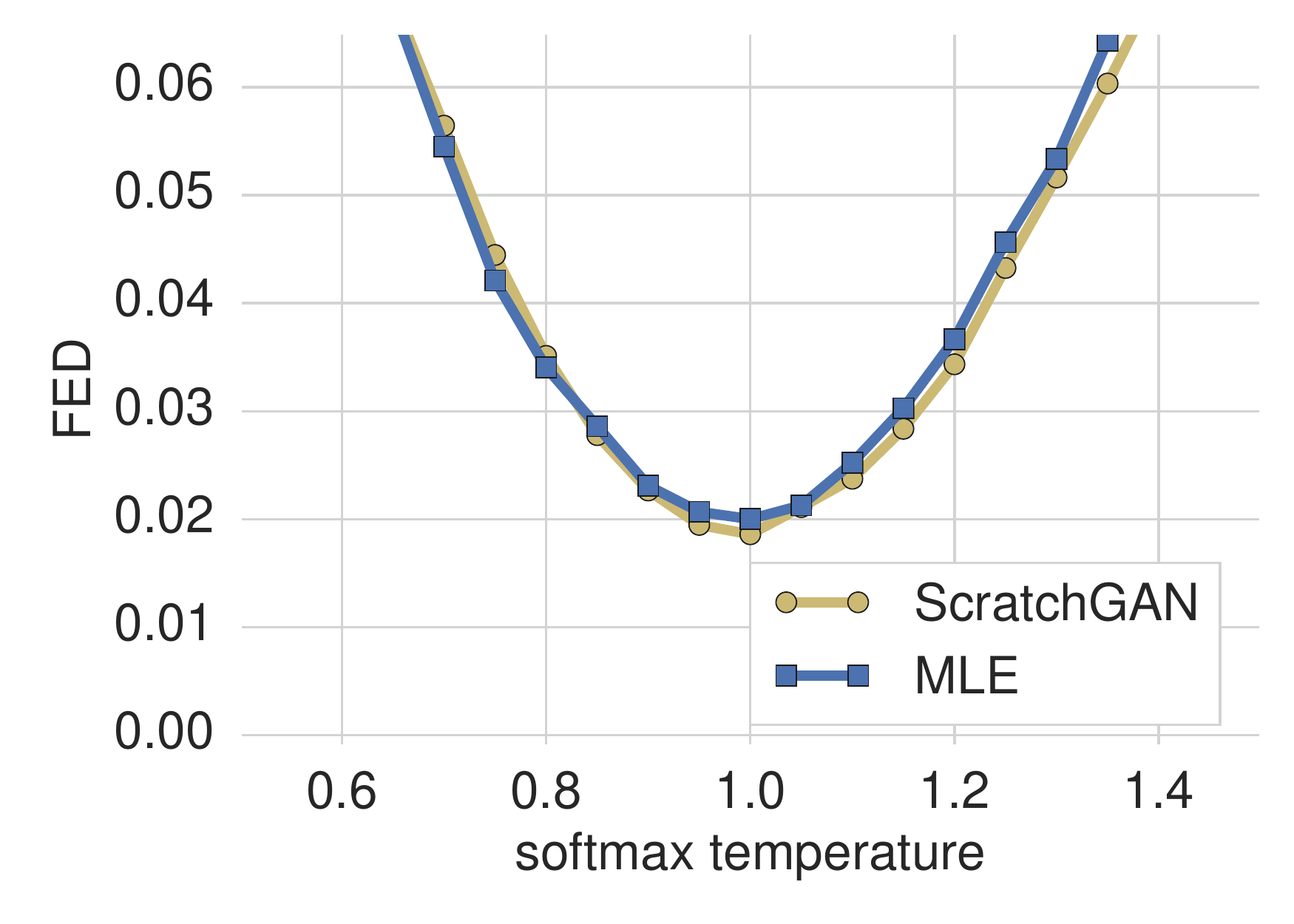}
  \captionsetup{justification=centering,margin=0cm}
  \caption{FED vs softmax temperature. }
   \label{fig:fd_vs_temp}
\end{subfigure}
\begin{subfigure}{.44\textwidth}
  \vspace{1mm}
  \centering
  \captionsetup{justification=raggedright,margin=0cm}
  \includegraphics[width=\linewidth]{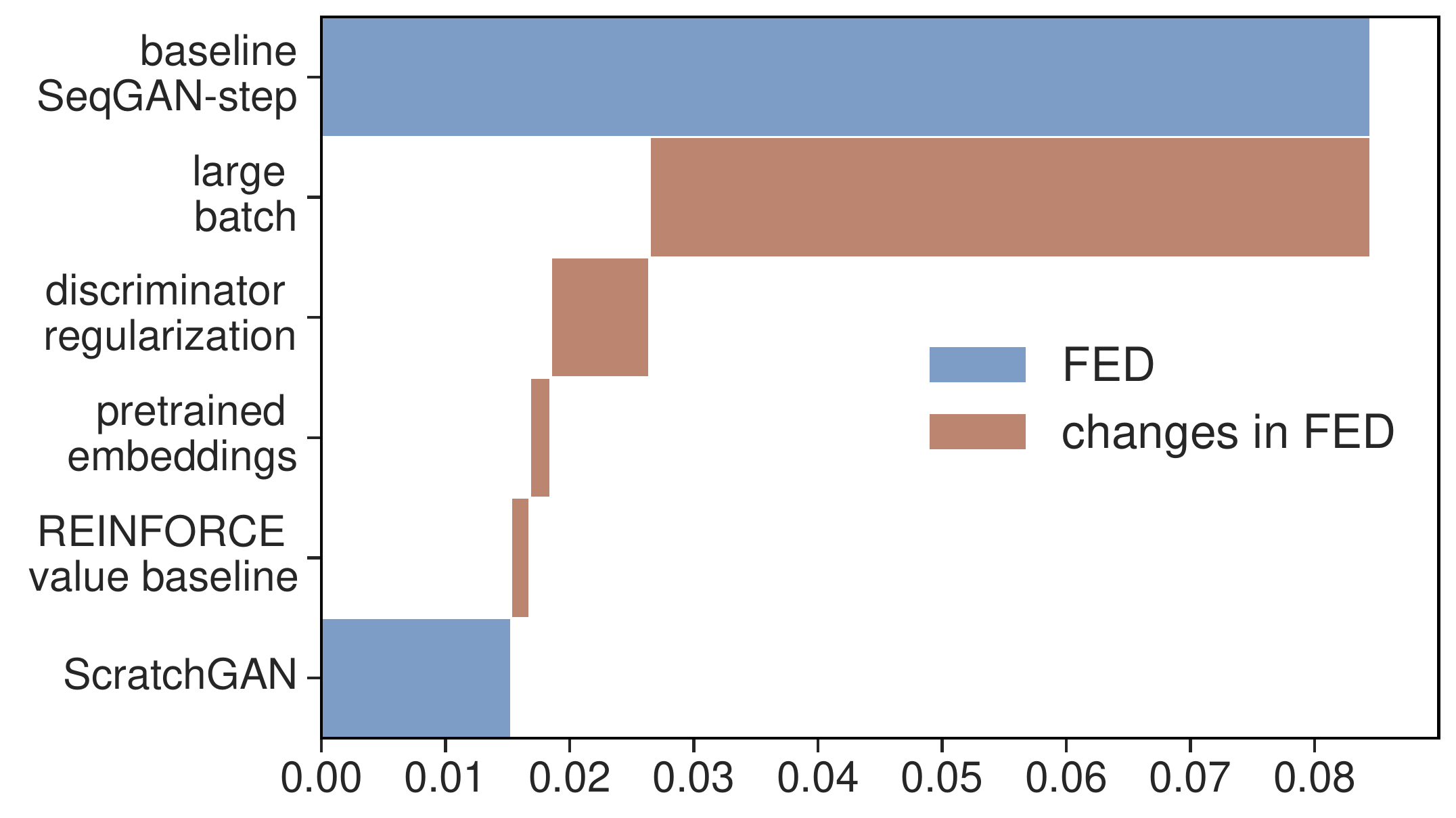}
  \caption{~\ourgan~ablation study.}
   \label{fig:ablation}
\end{subfigure}
\caption{FED scores. Lower is better. EMNLP2017 News results unless otherwise specified.}
\vspace{-3mm}
\end{figure}

\section{Experimental Results}
We use two datasets, EMNLP2017 News\footnote{\url{http://www.statmt.org/wmt17/}} and Wikitext-103 \cite{merity2016pointer}.
We use EMNLP2017 News to compare with prior work\citep{leakygan, fallingshort}
but note that this dataset has limitations: a small vocabulary (5.7k words), no out-of-vocabulary tokens, a sentence length limited to 50 tokens, and a size of only 300k sentences.
Wikitext-103 is a large scale dataset of almost 4 million sentences that captures more of the statistical properties of natural language and is a standard benchmark in language modeling~\citep{dauphin2016language,bai2018empirical}. For Wikitext-103 we use a vocabulary of 20k words.
In Wikitext-103 we remove sentences with less than 7 tokens or more than 100 tokens.
All our models are trained on individual sentences, using an NVIDIA P100 GPU.

In all our experiments, the baseline maximum likelihood trained language model is a dropout regularized LSTM. Model architectures, hyperparameters, regularization  and experimental procedures for the results below are detailed in
Appendix~\ref{app:exp_details}.
Samples from~\ourgan~can be seen in Appendix~\ref{app:samples}, alongside data and MLE samples.

\subsection{Quality and Diversity}
As suggested in ~\citet{fallingshort}, we measure the diversity-quality trade-off of different models by changing the softmax temperature at sampling time.
Reducing the softmax temperature below 1 results in higher quality but less diverse samples, while increasing it results in samples closer and closer to random.
Reducing the temperature for a language GANs is similar to the ``truncation trick" used in image GANs~\citep{biggan}.
We compute all metrics at different temperatures.

\ourgan~shows improved local consistency compared to existing language GANs and significantly reduces the gap between language GANs and the maximum likelihood language models. Figure \ref{fig:bleu_metrics} reports negative BLEU5 versus Self-BLEU5 metrics on EMNLP2017 News for \ourgan~and other language GANs, as reported in~\citet{fallingshort}.

\ourgan~improves over an MLE trained model on WikiText-103 according to FED, as shown in Figure~\ref{fig:wiki_fd}. This suggests that \ourgan~is more globally consistent and better captures semantic information.
Figure~\ref{fig:fd_vs_temp} shows the quality diversity trade-off as measured by FED as the softmax temperature changes. \ourgan~performs slightly better than the MLE model on this metric. This contrasts with the Language Model Score-Reverse Language Model scores shown in Figure~\ref{fig:wiki_lm_rlm}, which suggests that MLE samples are more diverse. Similar results on EMNLP2017 News are shown in Appendix~\ref{app:emnlp}.

\begin{figure*}[tb]
\begin{minipage}{.43\textwidth}
\centering
 \vspace{11mm}
  \begin{tabular}{c|c}
    \textbf{Model} & \textbf{World level perplexity} \\
      \midrule\midrule
    Random & 5725 \\
    \ourgan &  154 \\
    \textbf{MLE} & \textbf{42}  \\
  \end{tabular}
  \captionof{table}{EMNLP2017 News perplexity.}
  \label{tab:perplexity}
\end{minipage}%
\begin{minipage}{.48\textwidth}
\centerline{\includegraphics[width=0.8\columnwidth]{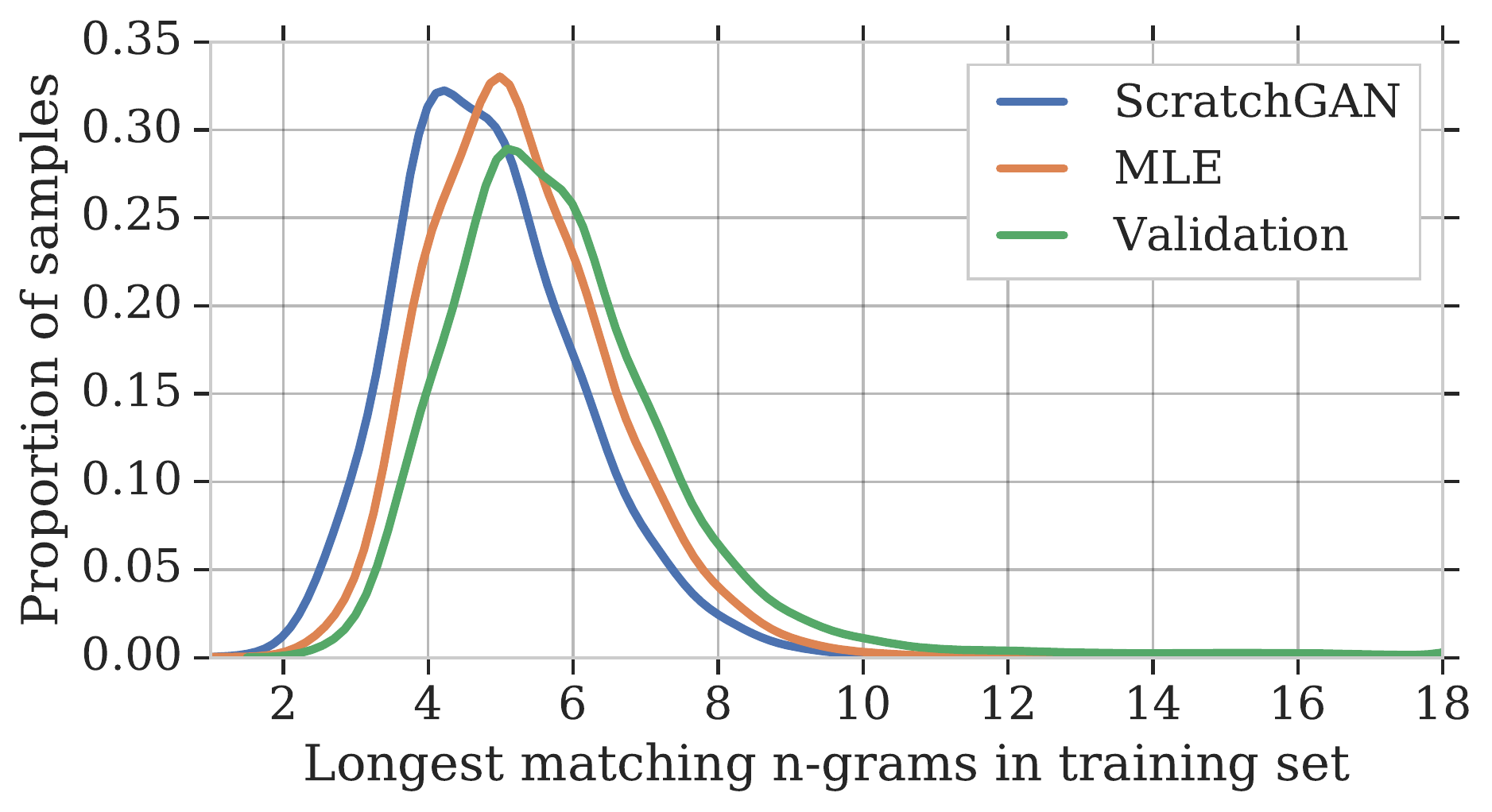}}
\captionsetup{justification=raggedright,margin=0cm}
\caption{Matching $n$-grams in EMNLP2017.}
\label{fig:ngram_dist}
\end{minipage}%
\vspace{-4mm}
\end{figure*}

Unlike image GANs,~\ourgan~learns an explicit model
of data, namely an autoregressive explicit model of language.
This allows us to compute model perplexities on validation data by feeding the model ground truth at each step.
We report~\ourgan~and MLE perplexities on EMNLP2017 News in Table~\ref{tab:perplexity}.
Evaluating perplexity favors the MLE model,
which is trained to minimize perplexity and thus has
an incentive to spread mass around the data distribution to avoid being penalized for not explaining training instances~\citep{theis2015note},
unlike~\ourgan~which is penalized by the discriminator when deviating from the data manifold and thus favors quality over diversity.
Improving sample diversity, together with avoiding underfitting by improving  grammatical and local consistency are required in order to further decrease
the perplexity of~\ourgan~to match that of MLE models.

Our diversity and quality evaluation across multiple metrics shows that compared to the MLE model,~\ourgan~trades off local consistency to achieve slightly better global consistency.

\subsection{Nearest Neighbors}
\label{sec:nn}
A common criticism of GAN models is that they produce realistic samples by overfitting to the training set, e.g. by copying text snippets.
 For a selection of \ourgan~samples we find and present the nearest neighbors present in the training set.
We consider two similarity measures, a 3-gram cosine similarity --- to capture copied word sequences,  and a cosine similarity from embeddings produced by the Universal Sentence Encoder ---to capture semantically similar sentences.
In Table~\ref{tab:nearest_neighbours} in Appendix~\ref{app:nearest_neighbours} we display a selection of four random samples and the corresponding top three closest training set sentences with respect to each similarity measure, and see the training text snippets have a mild thematic correspondence but have distinct phrasing and meaning. Additionally we perform a quantitive analysis over the full set of samples; we also compare the longest matching $n$-grams between text from the training set and (a) \ourgan~samples, (b) MLE samples, and (c) text from the validation set. In Figure~\ref{fig:ngram_dist} we see fewer \ourgan~samples with long matching n-grams ($n \ge 5$) in comparison with MLE samples and text from the validation set. We conclude the generator is producing genuinely novel sentences, although they are not always grammatically or thematically consistent.

\subsection{Ablation Study and SeqGAN-step comparison}
We show the relative importance of individual features of~\ourgan~with an ablation study in Figure~\ref{fig:ablation}.
We successively add all elements that appear important to~\ourgan~performance, namely large batch size, discriminator regularization ($L_2$ weight decay, dropout, and layer normalization), pre-trained embeddings, and a value baseline for REINFORCE. The increase in batch size results in the most significant performance boost, due to the reduction in gradient variance and stabilizing effect on adversarial dynamics. Discriminator regularization also leads to substantial performance gains, as it ensures the discriminator is not memorizing the training data and thus is providing a smoother learning signal for the generator.

The baseline model in Figure~\ref{fig:ablation} is a SeqGAN-step like model~\citep{seqgan} without pretraining. To highlight the improvement of~\ourgan~compared to prior work, we show in Table~\ref{tab:ablation} the FED difference between the two models.

\subsection{Training Stability}
Despite the high variance of REINFORCE gradients and the often unstable GAN training dynamics,
our training procedure is very stable, due to the use of large batch sizes and chosen reward structure.
Table~\ref{tab:fd_variance} reports the FED scores for~\ourgan~models trained with hyperparameters from a large volume in hyper-parameter space as well as across 50 random seeds.
The low variance across hyperparameters shows that~\ourgan~is not sensitive to changes in learning rate, REINFORCE discount factor, regularization or LSTM feature sizes, as long as these are kept in a reasonable range. The full hyperparameter sweep performed to obtain the variance estimates is described in Appendix~\ref{app:fd_variance_hyper}.
When we fixed hyperparameters and repeated an experiment across 50 seeds,
we obtained very similar FED score; no divergence or mode collapse occurred in any of the 50 runs.
For WikiText-103, the results are similar (0.055 $\pm$ 0.003).

\begin{table}[t]
  \begin{minipage}{.4\columnwidth}
    \centering
    \caption{FED on EMNLP2017 News.}
      \begin{tabular}{c|c}
        \textbf{Model}   & \textbf{FED}   \\ \midrule\midrule
        SeqGAN-step (no pretraining)   & 0.084 \\
        ScratchGAN              & \textbf{0.015} \\
      \end{tabular}
    \vspace{3mm}
      \label{tab:ablation}
  \end{minipage}
\hspace{5mm}
\begin{minipage}{0.55\columnwidth}
 \captionsetup{justification=raggedright}
\caption{FED sensitivity on EMNLP2017 News.}
  \centering
  \begin{tabular}{c|c}
    \textbf{Variation} & \textbf{FED} \\
      \midrule\midrule
    Hyperparameters &  0.021 $\pm$ 0.0056\\
    Seeds (best hypers) & 0.018 $\pm$ 0.0008\\
  \end{tabular}
\vspace{3mm}
    \label{tab:fd_variance}
\end{minipage}
\vspace{-8mm}
\end{table}

\section{Related Work}

Our work expands on the prior work of discrete language GANs, which opened up the avenues to this line of research.
Methods which use discrete data have proven to be more successful than methods using continuous relaxations~\citep{semeniuta2018accurate}, but face their own challenges, such as finding the right reward structure and reducing gradient variance.
Previously proposed solutions include: receiving dense rewards via Monte Carlo Search \citep{leakygan, seqgan, rankgan} or a recurrent discriminator~\citep{maskgan, semeniuta2018accurate}, leaking information from the discriminator to the generator \citep{leakygan}, using actor critic methods to reduce variance \citep{maskgan}, using ranking or moment matching to provide a richer learning signal \citep{rankgan, textgan} and curriculum learning~\citep{maskgan}. Despite alleviating problems somewhat, all of the above methods require pre-training, sometimes together with teacher forcing \citep{maligan} or interleaved supervised and adversarial training \citep{leakygan}.

\citet{relgan} recently showed that language GANs can benefit from complex architectures such as Relation Networks~\citep{relationnets}. Their RelGAN model can achieve better than random results without supervised pre-training, but still requires pre-training to achieve results comparable to MLE models.

\citet{press2017language} is perhaps the closest to our work: they train a character level GAN without pre-training. Unlike~\citet{press2017language}, \ourgan~is a word level model and does not require teacher helping, curriculum learning or continuous relaxations during training.
 Importantly, we have performed an extensive evaluation to quantify the performance of~\ourgan, as well as measured overfitting using multiple metrics, beyond $4$-gram matching.

By learning reward signals through the use of discriminators,
our work is in line with recent imitation learning work~\citep{ho2016generative},
as well as training non-differentiable generators~\citep{spiral}.

\section{Discussion}

Existing language GANs use maximum likelihood pretraining to minimize
adversarial training challenges, such as
unstable training dynamics and high variance gradient estimation.  However, they
have shown little to no performance improvements over traditional language models, likely due to
constraining the set of possible solutions to be close
to those found by maximum likelihood.
We have shown that large batch sizes, dense rewards and discriminator regularization remove the need for maximum likelihood pre-training in language GANs. To the best of our knowledge, we are the first to use \textit{Generative Adversarial Networks to train word-level language models successfully from scratch}.
Removing the need for maximum likelihood pretraining in language GANs opens up a new avenue of language modeling research, with
future work exploring GANs with one-shot feed-forward generators and
specialized discriminators which distinguish different features of language, such as semantics and syntax,
 local and global structure. Borrowing from the success of GANs for image generation~\citep{biggan}, another promising avenue is to use powerful neural network architectures~\citep{transformer,relationnets} to improve~\ourgan.

We have measured the quality and diversity of~\ourgan~samples using BLEU metrics, Fr\`echet distance, and language model scores.
None of these metrics is sufficient to evaluate language generation: we have shown that BLEU metrics only capture local consistency; language model scores do not capture semantic similarity; and that while embedding based Fr\`echet distance is a promising global consistency metric it is sensitive to sentence length. Until new ways to assess language generation are developed, current metrics need to be used together to compare models.

\section{Acknowledgments}

We would like to thank Chris Dyer, Oriol Vinyals, Karen Simonyan, Ali Eslami, David Warde-Farley, Siddhant Jayakumar and William Fedus for thoughtful discussions.

\clearpage
\balance
\bibliographystyle{unsrtnat}
\bibliography{main}%

\begin{thebibliography}{58}
\providecommand{\natexlab}[1]{#1}
\providecommand{\url}[1]{\texttt{#1}}
\expandafter\ifx\csname urlstyle\endcsname\relax
  \providecommand{\doi}[1]{doi: #1}\else
  \providecommand{\doi}{doi: \begingroup \urlstyle{rm}\Url}\fi

\bibitem[Wu et~al.(2016)Wu, Schuster, Chen, Le, Norouzi, Macherey, Krikun, Cao,
  Gao, Macherey, et~al.]{mt}
Yonghui Wu, Mike Schuster, Zhifeng Chen, Quoc~V Le, Mohammad Norouzi, Wolfgang
  Macherey, Maxim Krikun, Yuan Cao, Qin Gao, Klaus Macherey, et~al.
\newblock Google's neural machine translation system: Bridging the gap between
  human and machine translation.
\newblock \emph{arXiv preprint arXiv:1609.08144}, 2016.

\bibitem[Lample et~al.(2017)Lample, Conneau, Denoyer, and Ranzato]{unsupmt}
Guillaume Lample, Alexis Conneau, Ludovic Denoyer, and Marc'Aurelio Ranzato.
\newblock Unsupervised machine translation using monolingual corpora only.
\newblock \emph{arXiv preprint arXiv:1711.00043}, 2017.

\bibitem[Li et~al.(2016)Li, Monroe, Ritter, Galley, Gao, and Jurafsky]{dg}
Jiwei Li, Will Monroe, Alan Ritter, Michel Galley, Jianfeng Gao, and Dan
  Jurafsky.
\newblock Deep reinforcement learning for dialogue generation.
\newblock \emph{arXiv preprint arXiv:1606.01541}, 2016.

\bibitem[Allahyari et~al.(2017)Allahyari, Pouriyeh, Assefi, Safaei, Trippe,
  Gutierrez, and Kochut]{summarization}
Mehdi Allahyari, Seyedamin Pouriyeh, Mehdi Assefi, Saeid Safaei, Elizabeth~D
  Trippe, Juan~B Gutierrez, and Krys Kochut.
\newblock Text summarization techniques: a brief survey.
\newblock \emph{arXiv preprint arXiv:1707.02268}, 2017.

\bibitem[Vaswani et~al.(2017)Vaswani, Shazeer, Parmar, Uszkoreit, Jones, Gomez,
  Kaiser, and Polosukhin]{transformer}
Ashish Vaswani, Noam Shazeer, Niki Parmar, Jakob Uszkoreit, Llion Jones,
  Aidan~N Gomez, Lukasz Kaiser, and Illia Polosukhin.
\newblock Attention is all you need.
\newblock In \emph{Advances in Neural Information Processing Systems}, pages
  5998--6008, 2017.

\bibitem[Jozefowicz et~al.(2016)Jozefowicz, Vinyals, Schuster, Shazeer, and
  Wu]{jozefowicz2016exploring}
Rafal Jozefowicz, Oriol Vinyals, Mike Schuster, Noam Shazeer, and Yonghui Wu.
\newblock Exploring the limits of language modeling.
\newblock \emph{arXiv preprint arXiv:1602.02410}, 2016.

\bibitem[Radford et~al.(2019)Radford, Wu, Child, Luan, Amodei, and
  Sutskever]{radford2019language}
Alec Radford, Jeffrey Wu, Rewon Child, David Luan, Dario Amodei, and Ilya
  Sutskever.
\newblock Language models are unsupervised multitask learners.
\newblock \emph{OpenAI Blog}, 1:\penalty0 8, 2019.

\bibitem[Gu et~al.(2017)Gu, Bradbury, Xiong, Li, and Socher]{gu2017non}
Jiatao Gu, James Bradbury, Caiming Xiong, Victor~OK Li, and Richard Socher.
\newblock Non-autoregressive neural machine translation.
\newblock \emph{arXiv preprint arXiv:1711.02281}, 2017.

\bibitem[Bengio et~al.(2015)Bengio, Vinyals, Jaitly, and
  Shazeer]{bengio2015scheduled}
Samy Bengio, Oriol Vinyals, Navdeep Jaitly, and Noam Shazeer.
\newblock Scheduled sampling for sequence prediction with recurrent neural
  networks.
\newblock In \emph{Advances in Neural Information Processing Systems}, pages
  1171--1179, 2015.

\bibitem[Holtzman et~al.(2019)Holtzman, Buys, Forbes, and
  Choi]{nucleus_sampling}
Ari Holtzman, Jan Buys, Maxwell Forbes, and Yejin Choi.
\newblock The curious case of neural text degeneration.
\newblock \emph{arXiv preprint arXiv:1904.09751}, 2019.

\bibitem[Husz{\'a}r(2015)]{huszar2015not}
Ferenc Husz{\'a}r.
\newblock How (not) to train your generative model: Scheduled sampling,
  likelihood, adversary?
\newblock \emph{arXiv preprint arXiv:1511.05101}, 2015.

\bibitem[Goodfellow et~al.(2014)Goodfellow, Pouget-Abadie, Mirza, Xu,
  Warde-Farley, Ozair, Courville, and Bengio]{gan}
Ian Goodfellow, Jean Pouget-Abadie, Mehdi Mirza, Bing Xu, David Warde-Farley,
  Sherjil Ozair, Aaron Courville, and Yoshua Bengio.
\newblock Generative adversarial nets.
\newblock In \emph{NIPS}, 2014.

\bibitem[Gulrajani et~al.(2017)Gulrajani, Ahmed, Arjovsky, Dumoulin, and
  Courville]{improvedwgan}
Ishaan Gulrajani, Faruk Ahmed, Martin Arjovsky, Vincent Dumoulin, and Aaron
  Courville.
\newblock {Improved training of Wasserstein GANs}.
\newblock In \emph{NIPS}, 2017.

\bibitem[Yu et~al.(2017)Yu, Zhang, Wang, and Yu]{seqgan}
Lantao Yu, Weinan Zhang, Jun Wang, and Yong Yu.
\newblock Seqgan: Sequence generative adversarial nets with policy gradient.
\newblock 2017.

\bibitem[Guo et~al.(2017)Guo, Lu, Cai, Zhang, Yu, and Wang]{leakygan}
Jiaxian Guo, Sidi Lu, Han Cai, Weinan Zhang, Yong Yu, and Jun Wang.
\newblock Long text generation via adversarial training with leaked
  information.
\newblock \emph{arXiv preprint arXiv:1709.08624}, 2017.

\bibitem[Zhang et~al.(2017)Zhang, Gan, Fan, Chen, Henao, Shen, and
  Carin]{textgan}
Yizhe Zhang, Zhe Gan, Kai Fan, Zhi Chen, Ricardo Henao, Dinghan Shen, and
  Lawrence Carin.
\newblock Adversarial feature matching for text generation.
\newblock \emph{arXiv preprint arXiv:1706.03850}, 2017.

\bibitem[Che et~al.(2017)Che, Li, Zhang, Hjelm, Li, Song, and Bengio]{maligan}
Tong Che, Yanran Li, Ruixiang Zhang, R~Devon Hjelm, Wenjie Li, Yangqiu Song,
  and Yoshua Bengio.
\newblock Maximum-likelihood augmented discrete generative adversarial
  networks.
\newblock \emph{arXiv preprint arXiv:1702.07983}, 2017.

\bibitem[Lin et~al.(2017)Lin, Li, He, Zhang, and Sun]{rankgan}
Kevin Lin, Dianqi Li, Xiaodong He, Zhengyou Zhang, and Ming-Ting Sun.
\newblock Adversarial ranking for language generation.
\newblock In \emph{Advances in Neural Information Processing Systems}, pages
  3155--3165, 2017.

\bibitem[Semeniuta et~al.(2018)Semeniuta, Severyn, and
  Gelly]{semeniuta2018accurate}
Stanislau Semeniuta, Aliaksei Severyn, and Sylvain Gelly.
\newblock On accurate evaluation of gans for language generation.
\newblock \emph{arXiv preprint arXiv:1806.04936}, 2018.

\bibitem[Caccia et~al.(2018)Caccia, Caccia, Fedus, Larochelle, Pineau, and
  Charlin]{fallingshort}
Massimo Caccia, Lucas Caccia, William Fedus, Hugo Larochelle, Joelle Pineau,
  and Laurent Charlin.
\newblock Language gans falling short.
\newblock \emph{CoRR}, abs/1811.02549, 2018.
\newblock URL \url{http://arxiv.org/abs/1811.02549}.

\bibitem[Tevet et~al.(2018)Tevet, Habib, Shwartz, and
  Berant]{tevet2018evaluating}
Guy Tevet, Gavriel Habib, Vered Shwartz, and Jonathan Berant.
\newblock Evaluating text gans as language models.
\newblock \emph{arXiv preprint arXiv:1810.12686}, 2018.

\bibitem[Zhu et~al.(2018)Zhu, Lu, Zheng, Guo, Zhang, Wang, and
  Yu]{zhu2018texygen}
Yaoming Zhu, Sidi Lu, Lei Zheng, Jiaxian Guo, Weinan Zhang, Jun Wang, and Yong
  Yu.
\newblock Texygen: A benchmarking platform for text generation models.
\newblock \emph{arXiv preprint arXiv:1802.01886}, 2018.

\bibitem[Hochreiter and Schmidhuber(1997)]{lstm}
Sepp Hochreiter and J{\"u}rgen Schmidhuber.
\newblock Long short-term memory.
\newblock \emph{Neural computation}, 9\penalty0 (8):\penalty0 1735--1780, 1997.

\bibitem[Chung et~al.(2014)Chung, Gulcehre, Cho, and Bengio]{gru}
Junyoung Chung, Caglar Gulcehre, KyungHyun Cho, and Yoshua Bengio.
\newblock Empirical evaluation of gated recurrent neural networks on sequence
  modeling.
\newblock \emph{arXiv preprint arXiv:1412.3555}, 2014.

\bibitem[Shannon(1951)]{shannon1951prediction}
Claude~E Shannon.
\newblock Prediction and entropy of printed english.
\newblock \emph{Bell system technical journal}, 30\penalty0 (1):\penalty0
  50--64, 1951.

\bibitem[Mikolov et~al.(2010)Mikolov, Karafi{\'a}t, Burget, {\v{C}}ernock{\`y},
  and Khudanpur]{mikolov2010recurrent}
Tom{\'a}{\v{s}} Mikolov, Martin Karafi{\'a}t, Luk{\'a}{\v{s}} Burget, Jan
  {\v{C}}ernock{\`y}, and Sanjeev Khudanpur.
\newblock Recurrent neural network based language model.
\newblock In \emph{Eleventh Annual Conference of the International Speech
  Communication Association}, 2010.

\bibitem[Arjovsky et~al.(2017)Arjovsky, Chintala, and Bottou]{wgan}
Martin Arjovsky, Soumith Chintala, and L{\'e}on Bottou.
\newblock Wasserstein {GAN}.
\newblock In \emph{ICML}, 2017.

\bibitem[Mao et~al.(2016)Mao, Li, Xie, Lau, Wang, and Smolley]{lsgan}
Xudong Mao, Qing Li, Haoran Xie, Raymond~YK Lau, Zhen Wang, and Stephen~Paul
  Smolley.
\newblock Least squares generative adversarial networks.
\newblock \emph{arXiv preprint ArXiv:1611.04076}, 2016.

\bibitem[Jolicoeur-Martineau(2018)]{relativisticgan}
Alexia Jolicoeur-Martineau.
\newblock The relativistic discriminator: a key element missing from standard
  gan.
\newblock \emph{arXiv preprint arXiv:1807.00734}, 2018.

\bibitem[Mohamed and Lakshminarayanan(2016)]{implicitlearning}
Shakir Mohamed and Balaji Lakshminarayanan.
\newblock Learning in implicit generative models.
\newblock \emph{arXiv preprint arXiv:1610.03483}, 2016.

\bibitem[Williams(1992)]{reinforce}
Ronald~J Williams.
\newblock Simple statistical gradient-following algorithms for connectionist
  reinforcement learning.
\newblock \emph{Machine learning}, 8\penalty0 (3-4):\penalty0 229--256, 1992.

\bibitem[Maddison et~al.(2016)Maddison, Mnih, and Teh]{gumbel_softmax_concrete}
Chris~J Maddison, Andriy Mnih, and Yee~Whye Teh.
\newblock The concrete distribution: A continuous relaxation of discrete random
  variables.
\newblock \emph{arXiv preprint arXiv:1611.00712}, 2016.

\bibitem[Jang et~al.(2016)Jang, Gu, and Poole]{gumbel_softmax}
Eric Jang, Shixiang Gu, and Ben Poole.
\newblock Categorical reparameterization with gumbel-softmax.
\newblock \emph{arXiv preprint arXiv:1611.01144}, 2016.

\bibitem[Fedus et~al.(2017)Fedus, Rosca, Lakshminarayanan, Dai, Mohamed, and
  Goodfellow]{manypaths}
William Fedus, Mihaela Rosca, Balaji Lakshminarayanan, Andrew~M Dai, Shakir
  Mohamed, and Ian Goodfellow.
\newblock Many paths to equilibrium: Gans do not need to decrease adivergence
  at every step.
\newblock \emph{arXiv preprint arXiv:1710.08446}, 2017.

\bibitem[Fedus et~al.(2018)Fedus, Goodfellow, and Dai]{maskgan}
William Fedus, Ian Goodfellow, and Andrew~M Dai.
\newblock Maskgan: Better text generation via filling in the \_.
\newblock \emph{arXiv preprint arXiv:1801.07736}, 2018.

\bibitem[Nie et~al.(2019)Nie, Narodytska, and Patel]{relgan}
Weili Nie, Nina Narodytska, and Ankit Patel.
\newblock Rel{GAN}: Relational generative adversarial networks for text
  generation.
\newblock In \emph{International Conference on Learning Representations}, 2019.
\newblock URL \url{https://openreview.net/forum?id=rJedV3R5tm}.

\bibitem[Kannan and Vinyals(2017)]{orioladvdialogue}
Anjuli Kannan and Oriol Vinyals.
\newblock Adversarial evaluation of dialogue models.
\newblock \emph{arXiv preprint arXiv:1701.08198}, 2017.

\bibitem[Holtzman et~al.(2018)Holtzman, Buys, Forbes, Bosselut, Golub, and
  Choi]{holtzman2018learning}
Ari Holtzman, Jan Buys, Maxwell Forbes, Antoine Bosselut, David Golub, and
  Yejin Choi.
\newblock Learning to write with cooperative discriminators.
\newblock \emph{arXiv preprint arXiv:1805.06087}, 2018.

\bibitem[Sutton and Barto(2018)]{rlbook}
Richard~S Sutton and Andrew~G Barto.
\newblock \emph{Reinforcement learning: An introduction}.
\newblock 2018.

\bibitem[Pennington et~al.(2014)Pennington, Socher, and Manning]{glove}
Jeffrey Pennington, Richard Socher, and Christopher Manning.
\newblock Glove: Global vectors for word representation.
\newblock In \emph{Proceedings of the 2014 conference on empirical methods in
  natural language processing (EMNLP)}, pages 1532--1543, 2014.

\bibitem[Ba et~al.(2016)Ba, Kiros, and Hinton]{layernorm}
Jimmy~Lei Ba, Jamie~Ryan Kiros, and Geoffrey~E Hinton.
\newblock Layer normalization.
\newblock \emph{arXiv preprint arXiv:1607.06450}, 2016.

\bibitem[Srivastava et~al.(2014)Srivastava, Hinton, Krizhevsky, Sutskever, and
  Salakhutdinov]{dropout}
Nitish Srivastava, Geoffrey Hinton, Alex Krizhevsky, Ilya Sutskever, and Ruslan
  Salakhutdinov.
\newblock Dropout: a simple way to prevent neural networks from overfitting.
\newblock \emph{The Journal of Machine Learning Research}, 15\penalty0
  (1):\penalty0 1929--1958, 2014.

\bibitem[Brock et~al.(2018)Brock, Donahue, and Simonyan]{biggan}
Andrew Brock, Jeff Donahue, and Karen Simonyan.
\newblock Large scale gan training for high fidelity natural image synthesis.
\newblock \emph{arXiv preprint arXiv:1809.11096}, 2018.

\bibitem[Miyato et~al.(2018)Miyato, Kataoka, Koyama, and Yoshida]{spectralgan}
Takeru Miyato, Toshiki Kataoka, Masanori Koyama, and Yuichi Yoshida.
\newblock Spectral normalization for generative adversarial networks.
\newblock \emph{arXiv preprint arXiv:1802.05957}, 2018.

\bibitem[Papineni et~al.(2002)Papineni, Roukos, Ward, and
  Zhu]{papineni2002bleu}
Kishore Papineni, Salim Roukos, Todd Ward, and Wei-Jing Zhu.
\newblock Bleu: a method for automatic evaluation of machine translation.
\newblock In \emph{Proceedings of the 40th annual meeting on association for
  computational linguistics}, pages 311--318. Association for Computational
  Linguistics, 2002.

\bibitem[Reiter(2018)]{bleuissues}
Ehud Reiter.
\newblock A structured review of the validity of bleu.
\newblock \emph{Computational Linguistics}, pages 1--12, 2018.

\bibitem[Kneser and Ney(1995)]{kneser1995improved}
Reinhard Kneser and Hermann Ney.
\newblock Improved backing-off for m-gram language modeling.
\newblock In \emph{icassp}, volume~1, page 181e4, 1995.

\bibitem[Heusel et~al.(2017)Heusel, Ramsauer, Unterthiner, Nessler, and
  Hochreiter]{fid}
Martin Heusel, Hubert Ramsauer, Thomas Unterthiner, Bernhard Nessler, and Sepp
  Hochreiter.
\newblock Gans trained by a two time-scale update rule converge to a local nash
  equilibrium.
\newblock \emph{arXiv preprint arXiv:1706.08500}, 2017.

\bibitem[Cer et~al.(2018)Cer, Yang, Kong, Hua, Limtiaco, John, Constant,
  Guajardo-Cespedes, Yuan, Tar, et~al.]{cer2018universal}
Daniel Cer, Yinfei Yang, Sheng-yi Kong, Nan Hua, Nicole Limtiaco, Rhomni~St
  John, Noah Constant, Mario Guajardo-Cespedes, Steve Yuan, Chris Tar, et~al.
\newblock Universal sentence encoder.
\newblock \emph{arXiv preprint arXiv:1803.11175}, 2018.

\bibitem[Merity et~al.(2016)Merity, Xiong, Bradbury, and
  Socher]{merity2016pointer}
Stephen Merity, Caiming Xiong, James Bradbury, and Richard Socher.
\newblock Pointer sentinel mixture models.
\newblock \emph{arXiv preprint arXiv:1609.07843}, 2016.

\bibitem[Dauphin et~al.(2016)Dauphin, Fan, Auli, and
  Grangier]{dauphin2016language}
Yann~N Dauphin, Angela Fan, Michael Auli, and David Grangier.
\newblock Language modeling with gated convolutional networks.
\newblock \emph{arXiv preprint arXiv:1612.08083}, 2016.

\bibitem[Bai et~al.(2018)Bai, Kolter, and Koltun]{bai2018empirical}
Shaojie Bai, J~Zico Kolter, and Vladlen Koltun.
\newblock An empirical evaluation of generic convolutional and recurrent
  networks for sequence modeling.
\newblock \emph{arXiv preprint arXiv:1803.01271}, 2018.

\bibitem[Theis et~al.(2015)Theis, Oord, and Bethge]{theis2015note}
Lucas Theis, A{\"a}ron van~den Oord, and Matthias Bethge.
\newblock A note on the evaluation of generative models.
\newblock \emph{arXiv preprint arXiv:1511.01844}, 2015.

\bibitem[Santoro et~al.(2017)Santoro, Raposo, Barrett, Malinowski, Pascanu,
  Battaglia, and Lillicrap]{relationnets}
Adam Santoro, David Raposo, David~G Barrett, Mateusz Malinowski, Razvan
  Pascanu, Peter Battaglia, and Timothy Lillicrap.
\newblock A simple neural network module for relational reasoning.
\newblock In \emph{Advances in neural information processing systems}, pages
  4967--4976, 2017.

\bibitem[Press et~al.(2017)Press, Bar, Bogin, Berant, and
  Wolf]{press2017language}
Ofir Press, Amir Bar, Ben Bogin, Jonathan Berant, and Lior Wolf.
\newblock Language generation with recurrent generative adversarial networks
  without pre-training.
\newblock \emph{arXiv preprint arXiv:1706.01399}, 2017.

\bibitem[Ho and Ermon(2016)]{ho2016generative}
Jonathan Ho and Stefano Ermon.
\newblock Generative adversarial imitation learning.
\newblock In \emph{Advances in Neural Information Processing Systems}, pages
  4565--4573, 2016.

\bibitem[Ganin et~al.(2018)Ganin, Kulkarni, Babuschkin, Eslami, and
  Vinyals]{spiral}
Yaroslav Ganin, Tejas Kulkarni, Igor Babuschkin, SM~Eslami, and Oriol Vinyals.
\newblock Synthesizing programs for images using reinforced adversarial
  learning.
\newblock \emph{arXiv preprint arXiv:1804.01118}, 2018.

\bibitem[Kingma and Ba(2014)]{kingma2014adam}
Diederik~P Kingma and Jimmy Ba.
\newblock Adam: A method for stochastic optimization.
\newblock \emph{arXiv preprint arXiv:1412.6980}, 2014.

\end{thebibliography}

\clearpage

\appendix
{\centerline{\Large{\textbf{Supplementary material}}}}

\section{Fr\'echet Embedding Distance and Language model scores on EMNLP2017 News}
\label{app:emnlp}

\begin{figure}[h]
\centering
\caption{EMNLP2017 News results.}
\begin{subfigure}{.45\textwidth}
  \centering
  \captionsetup{justification=centering,margin=0cm}
  \includegraphics[width=\linewidth]{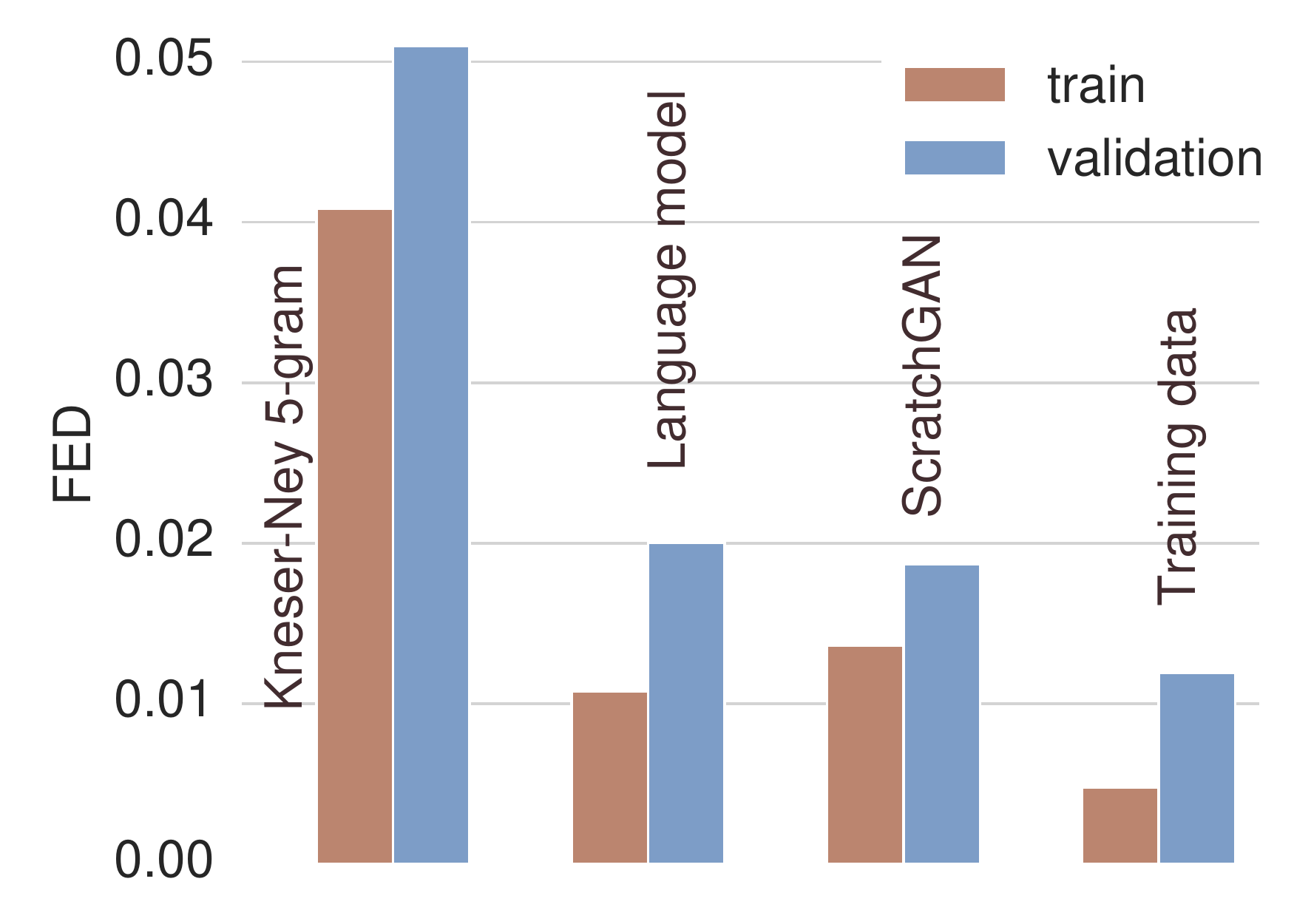}
  \caption{FED against training and validation data.}
  \label{fig:emnlp_fd}
\end{subfigure}%
\hspace{2mm}
\begin{subfigure}{.5\textwidth}
  \centering
  \includegraphics[width=0.78\linewidth]{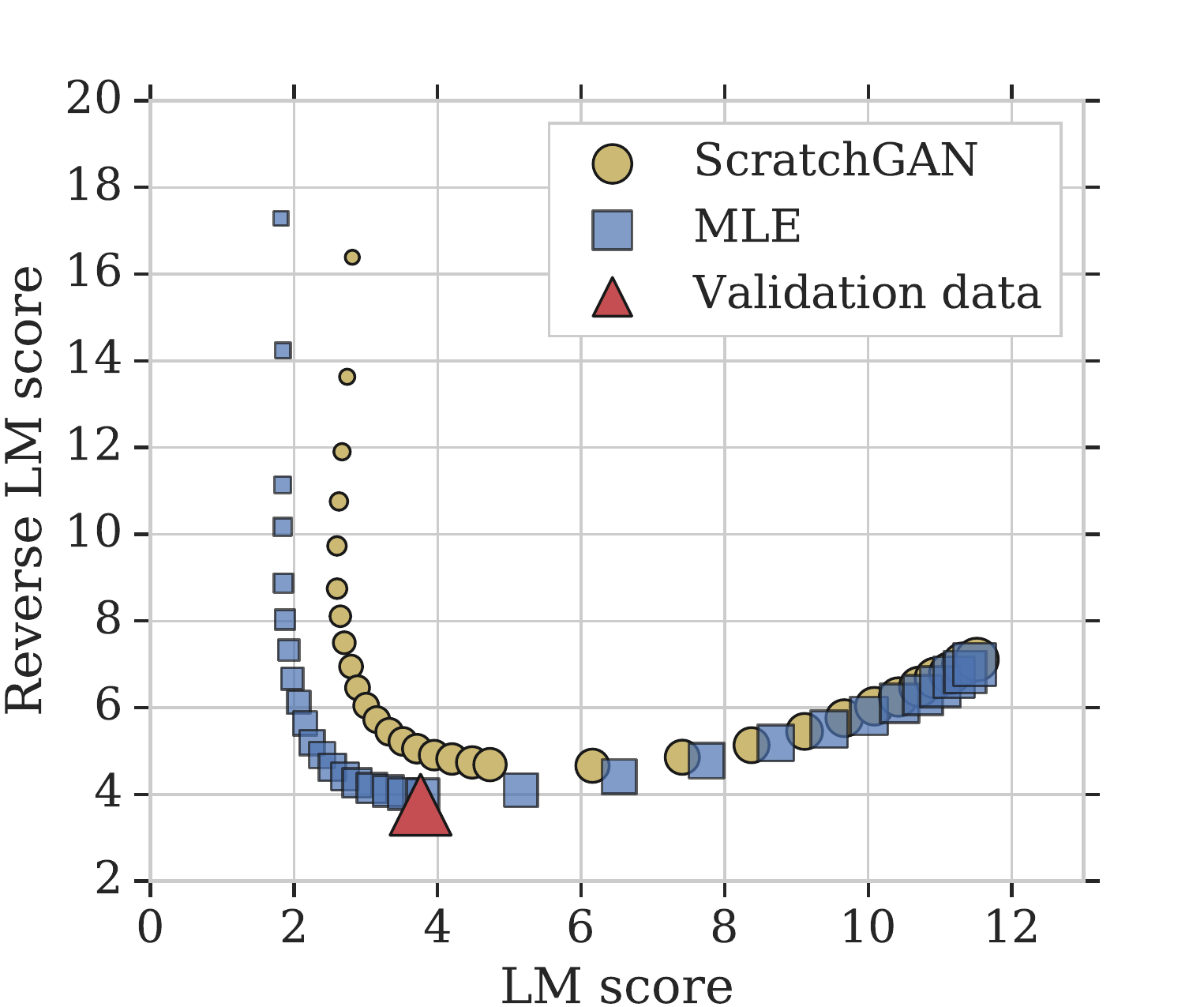}
  \caption{Language model scores.}
  \label{fig:emnlp_lm_rlm}
\end{subfigure}
\vspace{-3mm}
\end{figure}

On EMNLP2017 News, FED and LM/RLM results are similar to those on WikiText103, see Figure~\ref{fig:emnlp_fd} and Figure~\ref{fig:emnlp_lm_rlm}. Here we report the FED
against both the training and validation set, to assess model overfitting.
On this metric, we again notice that~\ourgan~performs better than the MLE model.

\section{Nearest Neighbors}
\label{app:nearest_neighbours}

In Table~\ref{tab:nearest_neighbours} we see for a selection of four random samples and the corresponding top three closest training set sentences with respect to each similarity measure, there is not a clear pattern of overfitting or training set repetition.

\begin{table*}[th]
\caption{EMNLP2017 News nearest neighbours to~\ourgan~samples. Similarity with respect to embedding cosine distance using the Universal Sentence Encoder, and with respect to 3-gram cosine distance. We see the GAN samples are not composed of cut-and-paste text snippets from the training set.}
\begin{center}
\small
\def\arraystretch{1.5}
\begin{tabular}{|l p{6cm}|l p{6cm}|}
\hline
USE & Nearest Neighbours & 3-gram & Nearest Neighbours \\
\hline
\hline
\multicolumn{4}{|l|}{Sample: \textit{A nice large part of Trump has to plan exactly what Pence would worth , for Trump to choose him }} \\
\multicolumn{4}{|l|}{\hspace{1em} \textit{strongly in Florida, where he can be 100 percent away.}} \\
0.77 & His name , of course , is Donald Trump , the billionaire businessman who leads most national polls for the Republican nomination . &
0.13 & It ' s like the situation in Florida , where he didn ' t pay taxes on his golf course . \\
0.75 & But to get there , Rubio believes he needs to cut significantly into Cruz ' s support in Iowa , a state dominated by social conservatives . &
0.12 & Donald Trump is spending his third straight day in Florida , where he ' s already made six campaign stops since Sunday . \\
0.72 & On the Republican side , the Iowa poll shows Ted Cruz leading Donald Trump by four points , but Trump has a 16 - point lead in New Hampshire .
& 0.10 & He has long been mentioned as a possible candidate for governor in Florida , where he has a home in Miami with his wife and four school - age children . \\
\hline
\hline
\multicolumn{4}{|l|}{Sample: \textit{I didn ' t know how to put him up to the floor among reporters Thursday or when he did what he said.}} \\
0.69 & Speaking at a news conference on Monday , he said : " Let me make clear that this is a great professional and a great person . & 0.25 & Her explanation for saying " I didn ' t glass her , I don ' t know why I ' m getting arrested " was said out of panic , I didn ' t know how to handle the situation . \\
0.67 & In a text message late Monday , he said he had not seen the court filing and could not comment on it . & 0.23 & I didn ' t know how to do it or who to talk to , so I had to create opportunities for myself . \\
0.59 & " We ' re not going to stand by any agent that has deliberately done the wrong thing , " he said . & 0.23 & I didn ' t know how to face it , but as soon as I ' d got through that it was OK . \\
\hline
\hline
\multicolumn{4}{|l|}{Sample: \textit{Paul have got a fine since the last 24 game , and it ' s just a nine - day mark .}} \\
0.50 & As he said after Monday night ' s game : " We know we have enough quality , it ' s not always the quality . & 0.21 & We ' ve been in this situation too many times , and it ' s a 60 - minute game , and it doesn ' t matter . \\
0.50 & The 26 - year - old from Brisbane was forced to come from behind to score an impressive 6 - 7 ( 5 - 7 ), 6 - 4 , 7 - 6 ( 8 - 6 ) win . & 0.21 & There are already plenty people fighting fire with fire , and it ' s just not helping anyone or anything . \\
0.48 & But he ' s had a very good start to this year and beat Roger to win Brisbane a couple of weeks ago . & 0.20 & We ' ve just got to move on , it ' s part of the game , and it ' s always going to happen , that kind of stuff  \\
\hline
\hline
\multicolumn{4}{|l|}{Sample: \textit{Such changes from the discussion and social support of more people living in the EU with less generous }} \\
\multicolumn{4}{|l|}{\textit{income and faith.}} \\
0.72 & The EU has promised Ankara three billion euros in aid if it does more to stop the flow of migrants headed for Europe . & 0.14 & There are nearly three - quarters of a million British people living in Spain and over two million living in the EU as a whole . \\
0.68 & Now , as Norway is not a member of the EU , it has no say over these or any other EU rules . &  0.1  & About 60 people living in the facility were moved to another part of the building for safety , according to authorities . \\
0.67 & We can ' t debate the UK ' s place in Europe ahead of an historic EU referendum without accurate statistics on this and other issues . & 0.1 & We ' d like to hear from people living in the country about what life as a Canadian is really like . \\
\hline
\end{tabular}
\end{center}
\label{tab:nearest_neighbours}
\end{table*}

\section{Negative results}
\label{app:neg_res}

Here we list some approaches that we tried but which proved unsuccessful or unnecessary:
\begin{itemize}
  \item Using a Wasserstein Loss on generator logits, with a straight-through gradient. This was unsuccessful.
  \item Using ensembles of discriminators and generators. The results are on par with those obtained by a single discriminator-generator pair.
  \item Training against past versions of generators/discriminators. Same as above.
  \item Using bi-directional discriminators. They can work but tend to over-fit and provide less useful feedback to the generator.
  \item Using several discriminators with different architectures, hoping to have the simple discriminators capture simple failure modes of the generators such as repeated words. It did not improve over single discriminator-generator pair.
  \item Training on small datasets such as Penn Tree Bank. The discriminator quickly over-fit to the training data. This issue could probably be solved with stronger regularization but we favoured larger datasets.
  \item Using a Hinge loss~\citep{spectralgan} on the discriminator. This did not improve over the cross-entropy loss.
  \item Using a hand-designed curriculum, where the generator is first trained against a simple discriminator, and later in training a more complex discriminator is substituted. This was unsuccessful. We suspect that adversarial training requires a difficult balance between discriminator quality and generator quality, which is difficult to reach when either component has been trained independently from the other.
  \item Varying significantly the number of discriminator steps per generator step, say 5 discriminator steps per generator step. This was unsuccessful.
 \item Looking at discriminator probabilities (check that $P(real)\approx1$ and $P(fake)\approx 0$) to evaluate training. Discriminator seems to be able to provide good gradient signal even when its predictions are not close to the targets, as long as its predictions on real data are distinct from its prediction on fake data.
 \item Using a population of discriminators to evaluate the quality of a generator, or conversely. This metric failed when the population as a whole is not making progress.
 \item Mapping all data to GloVe embeddings, and training a one-shot feed-forward generator to generate word embeddings directly, while discriminator receives word embeddings directly. This was unsucessful.
\end{itemize}

\section{Experimental details}
\label{app:exp_details}

We now provide the experimental details of our work.

\subsection{\ourgan~architectural details}

\textbf{Generator}\\
The core of the generator is an LSTM with tanh activation function and skip connections.
We use an embedding matrix which is the concatenation of a fixed pretrained GloVe embedding matrix of dimension $V\times 300$ where $V$ is the vocabulary size, and a learned embedding matrix of dimension $V \times M$ where $M$ depends on the dataset.
An embedding for the token at the previous time-step is looked up in the embedding matrix, and then
linearly projected using a learned matrix to the feature size of the LSTM.
This is the input to the LSTM.
The output of the LSTM is the concatenation of the hidden outputs of all layers.
This output is linearly projected using a learned matrix to the dimension of the embedding matrix.
We add a learned bias of dimension $V$ to obtain the logits over the vocabulary.
We apply a softmax operation to the logits to obtain a Categorical distribution and sample from it to generate the token for the current time-step.

\textbf{Discriminator}\\
The input to the discriminator is a sequence of tokens, coming either from the real data or the generator.
The core of the discriminator is an LSTM.
The discriminator uses its own embedding matrix, independent from the generator.
It has the same structure as the generator embedding matrix.
Dropout is applied to this embedding matrix.
An embedding for the token at the current time-step $t$ is looked up in the embedding matrix.
A fixed position embedding of dimension $8$, depending on $t$ (see \ref{app:pos_info}), is concatenated to the embedding.
As for the generator, the embedding is linearly projected using a learned matrix to the feature size of the LSTM.
This is the input to the LSTM.
The output of the LSTM is itself linearly projected to dimension $1$.
This scalar is passed through a sigmoid to obtain the discriminator probability $\disc(\vx_t)$.
The discriminator LSTM is regularized with layer normalization.
$L_2$ regularization is applied to all learned variables in the discriminator.

\textbf{Losses} \\
The discriminator is trained with the usual cross-entropy loss.
The generator is trained with a REINFORCE loss. The value baseline at training step $i$, denoted $b_i$, is computed as:
\begin{equation}
\label{eq:baseline_decay}
b_i = \lambda b_{i-1} + (1-\lambda)\bar{R}_i
\end{equation}
where $\bar{R}_i$ is the mean cumulative reward over all sequence timesteps and over the current batch at training step $i$. The generator loss at timestep $t$ and training step $i$ is then:
\begin{equation}
L^G_{ti} = - (R_t-b_i) \ln p_\theta(x_t)
\end{equation}
and the total generator loss to minimize at training step $i$ is $\sum_t L^G_{ti}$.

\textbf{Optimization} \\
Both generators and discriminators are trained with Adam~\citep{kingma2014adam}, with $\beta_1=0.5$ for both.
We perform one discriminator step per generator step.

\textbf{Data considerations} \\
The maximum sequence length for EMNLP2017 News is 50 timesteps.
The generator vocabulary also contains a special end of sequence token.
If the generator outputs the end of sequence token at any timestep the rest of the sequence is padded with spaces.
At timestep $0$ the input to the generator LSTM is a space character.
Generator and discriminator are both recurrent so time and space complexity of inference and training are linear in the sequence length.

\subsection{Sweeps and best hyperparameters}

To choose our best model, we sweep over the following hyperparameters:
\begin{itemize}[itemsep=0.0em]
  \item Discriminator learning rate.
  \item Generator learning rate.
  \item Discount factor $\gamma$.
  \item The number of discriminator updates per generator update.
  \item The LSTM feature size of the discriminator and generator.
  \item The number of layers for the generator.
  \item Batch size.
  \item Dropout rate for the discriminator.
  \item Trainable embedding size.
  \item Update frequency of baseline, $\lambda$.
\end{itemize}

The best hyperparameters for EMNLP2017 News are:
\begin{itemize}[itemsep=0.0em]
  \item Discriminator learning rate: $9.38\,10^{-3}$.
  \item Generator learning rate: $9.59\,10^{-5}$
  \item Discount factor $\gamma$: 0.23.
  \item The LSTM feature size of the discriminator and generator: 512 and 512.
  \item The number of layers for the generator: 2.
  \item Batch size: 512.
  \item Dropout rate for the discriminator embeddings: 0.1
  \item Trainable embedding size: 64.
  \item Update frequency of baseline, $\lambda$: 0.08.
\end{itemize}

The best hyperparameters for WikiText-103 News:
\begin{itemize}[itemsep=0.0em]
  \item Discriminator learning rate: $2.98\,10^{-3}$
  \item Generator learning rate: $1.67\,10^{-4}$
  \item Discount factor $\gamma$: 0.79.
  \item The LSTM feature size of the discriminator and generator: 256 and 256.
  \item The number of layers for the discriminator: 1.
  \item Batch size: 768.
  \item Dropout rate for the discriminator embeddings: 0.4.
  \item Trainable embedding size: 16.
  \item Update frequency of baseline, $\lambda$: 0.23.
\end{itemize}

\subsection{Training procedure}
For both datasets, we train for at least $100000$ generator training steps, saving the model every $1000$ steps,
and we select the model with the best FED against the validation data.
Each training run used approximately 4 Intel Skylake x86-64 CPUs at 2 GHz, 1 Nvidia Tesla V100 GPU, and 20 GB of RAM, for 1 to 5 days depending on the dataset.

\subsection{Language models}

The language models we compare to are LSTMs.
Interestingly, we found that smaller architectures are necessary for the LM compared to the GAN model, in order to avoid overfitting.
For the maximum likelihood language models,
we sweep over the size of the embedding layer, the feature size of the LSTM,
and the dropout rate used for the embedding layer. We choose the model
with the smallest validation perplexity.

For EMNLP2017 News, the MLE model used a LSTM feature size of 512, embedding size of 512, and embedding dropout rate of 0.2.

For WikiText-103, the MLE model used a LMST feature size of 3000, embedding size of 512, and embedding dropout rate of 0.3.

\subsection{Metrics}

FED and BLEU/Self-BLEU metrics on EMNLP2017 News are always computed with $10000$ samples.
On WikiText-103 FED is computed with $7869$ samples because this is the number of sentences in WikiText-103 validation data, after filtering outliers.

To compute the reverse language model scores at different softmax temperatures we used the same architecture as the best EMNLP2017 News. We trained a language model on 268590 model samples, and used it to score the validation data.

\subsection{Datasets}
Wikitext-103 is available at \url{https://s3.amazonaws.com/research.metamind.io/wikitext/wikitext-103-v1.zip}.
EMNLP2017News is available at \url{http://www.statmt.org/wmt17/} and a preprocessed version at \url{https://github.com/pclucas14/GansFallingShort/blob/master/real_data_experiments/data/news/}.

\section{Fr\'echet Embedding Distance sensitivity to sentence length}
\label{app:fd_sen_length}
\begin{figure}[h]
\centering
\caption{EMNLP2017 News results.}
\begin{subfigure}{.45\textwidth}
  \centering
  \centerline{\includegraphics[width=\columnwidth]{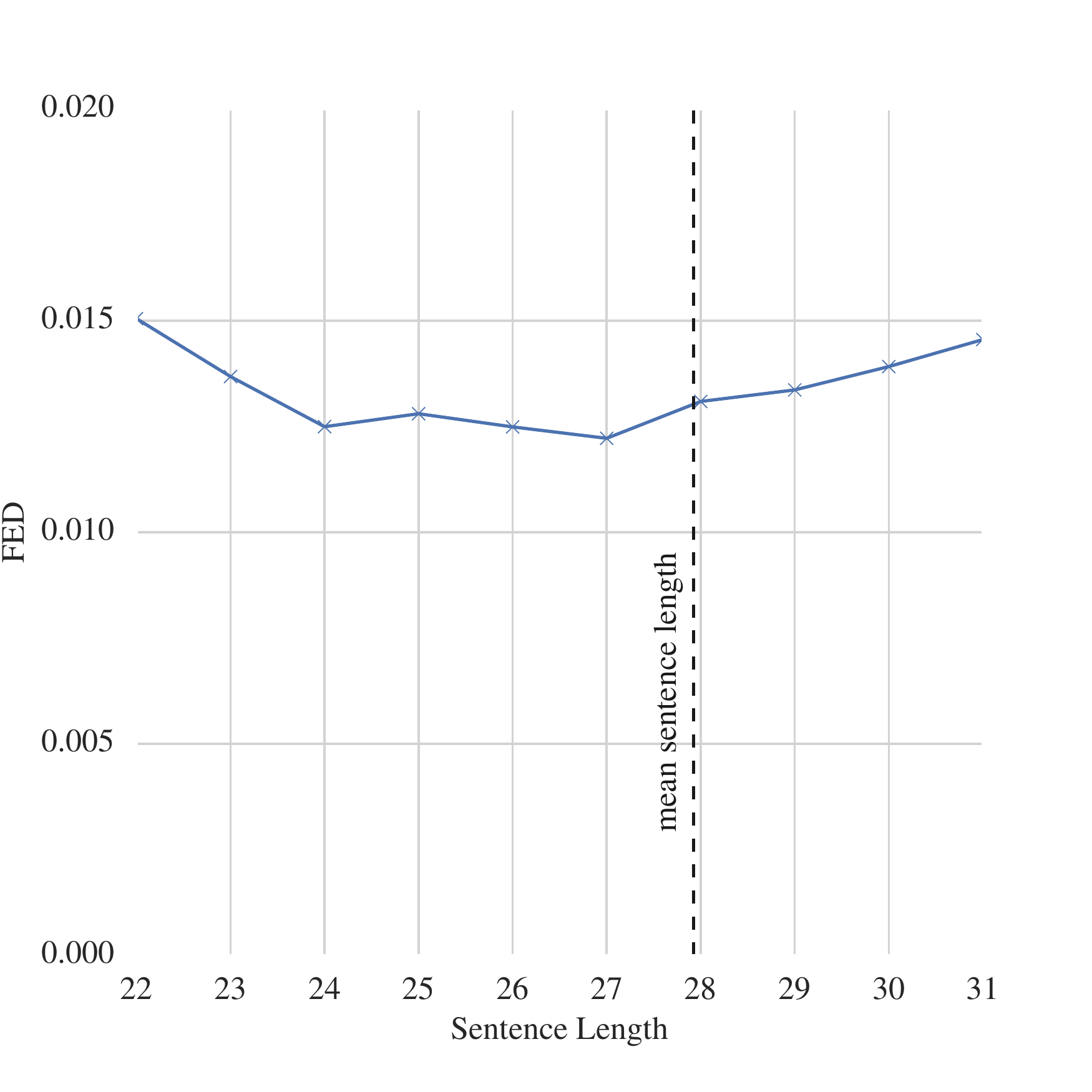}}
  \caption{FED vs sentence length.}
  \label{fig:fd_vs_length}
\end{subfigure}%
\hspace{2mm}
\begin{subfigure}{.5\textwidth}
  \centering
  \centerline{\includegraphics[width=\columnwidth]{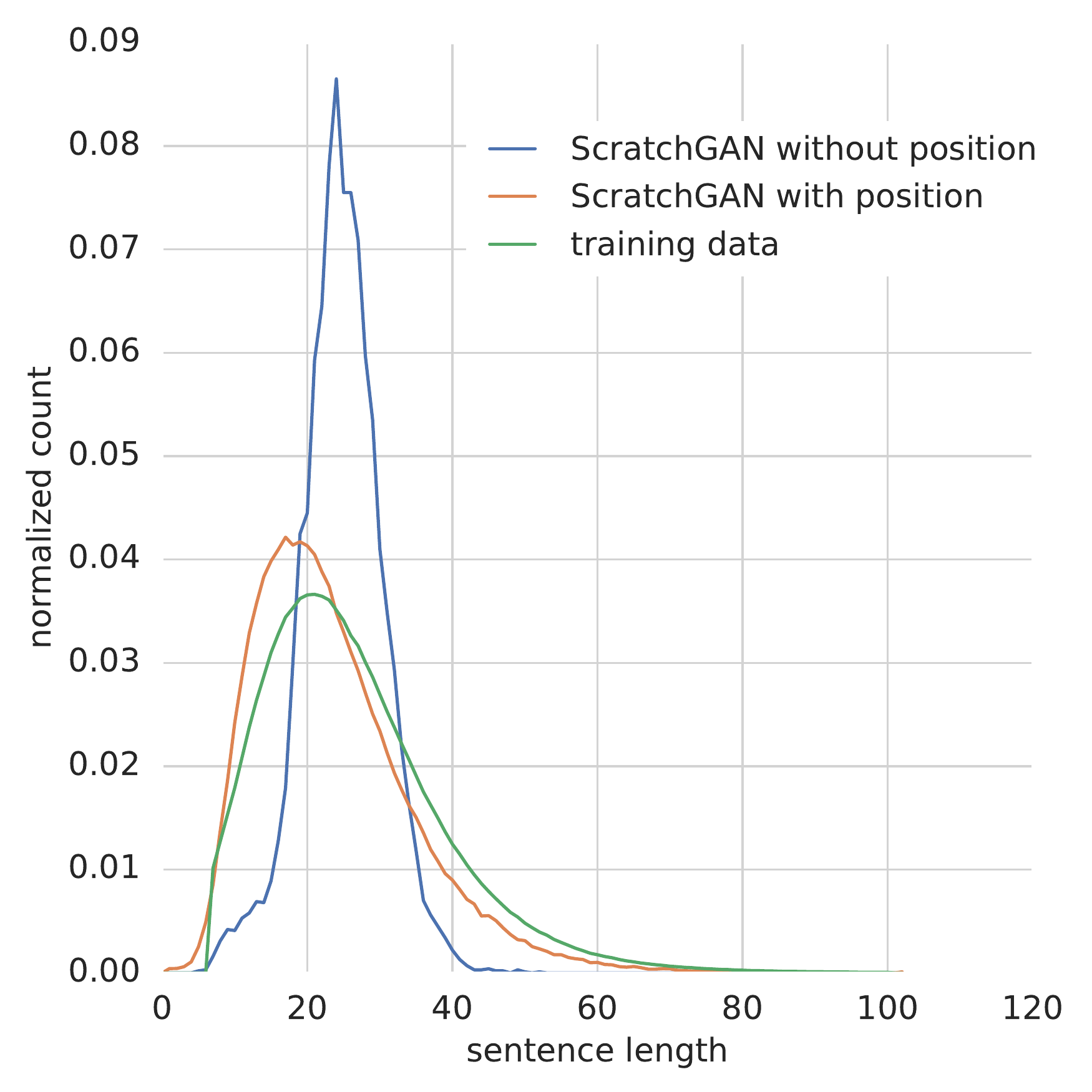}}
  \caption{Providing positional information to the discriminator helps the generator capture sentence length distribution correctly.}
  \label{fig:pos_info}
\end{subfigure}
\vspace{-3mm}
\end{figure}

We show that FED is slightly dependent on sentence length, highlighting a possible limitation of this metric.
For each sentence length, we randomly select a subset of 10k sentences from EMNLP2017 News training set conditioned on this sentence length, and we measure the FED between this subset and the 10k validation set.
We show the results in figure~\ref{fig:fd_vs_length}.
We see that there is a small dependence of FED on sentence length: FED seems to be worse for sentences that are significantly shorter or longer than the mean.

\section{Hyperparameter variance}
\label{app:fd_variance_hyper}

Here we clarify the definition of the subset of hyper-parameter space that we used to show the stability of our training procedure. All runs with hyper-parameters in the ranges defined below gave good results in our experiments as shown in Table~\ref{tab:fd_variance} in the main text.

\begin{itemize}
\item baseline decay ($\lambda$ in equation \ref{eq:baseline_decay} in appendix \ref{app:exp_details}) is in $[0, 1]$.
\item batch size is in $\{512, 768\}$
\item discriminator dropout is in $\{0.1, 0.2, 0.3, 0.4, 0.5\}$
\item discriminator LSTM feature size is in $\{256, 512, 1024\}$
\item discriminator learning rate is in $[9.2\,10^{-5}, 3.7\,10^{-2}]$
\item discriminator $L_2$ weight is in $\{0, 10^{-7}, 10^{-6}, 10^{-5}\}$
\item discriminator LSTM number of layers is in $\{1, 2\}$
\item number of discriminator updates per training step is in $\{1, 2\}$
\item discount factor in REINFORCE is in $[0, 1]$
\item generator LSTM feature size is in $\{256, 512\}$
\item generator learning rate is in $[8.4\,10^{-5}, 3.4\,10^{-4}]$
\item generator LSTM number of layers is in $\{1, 2\}$
\item number of generator updates per training step is in $\{1, 2\}$
\item dimension of trainable embeddings is in $\{16, 32, 64\}$
\end{itemize}

\section{Positional information provided to the discriminator}
\label{app:pos_info}

Here we discuss the importance of providing positional information to the discriminator.
In early experiments we noticed that the distribution of sentence length in the generator samples did not match the distribution of sentence length found in the real data.
In theory, we would expect a discriminator based on a LSTM to be able to easily spot samples that are significantly too short or long, and to provide that signal to the generator.
But in practice, the generator was biased towards avoiding short and long sentences.

We therefore provide the discriminator with explicit positional information, by concatenating a fix sinusoidal signal to the word embeddings used in the discriminator.
We choose 8 periods log-linearly spaced $(T_1, \dots, T_8)$ such that $T_1=2$ and $T_8$ is 4 times the maximum sentence length.
For the token $x_t$ at position $t$ in the sentence, the positional information is $p^i_t=\sin \bigl(2\pi \frac{t}{T_i}\bigr)$.
We concatenate this positional information to the word embedding for token $x_t$ in the discriminator before using it as input for the discriminator LSTM.

Figure~\ref{fig:pos_info} shows distributions of sentence length in samples of two GAN models, one with and one without this positional information. We compare these distributions against the reference distribution of sentence length in the training data. Even with positional information in the discriminator, the generator still seems slightly biased towards shorter sentences, compared to the training data. But the sentence length distribution is still a much better fit with positional information than without.

\section{Samples}
Training examples from both datasets can be found in Table~\ref{tab:data}.
Samples from our model, the maximum likelihood trained language model
and the $n$-gram model can be found in Tables
~\ref{tab:more_samples}
,~\ref{tab:mle_samples} and
~\ref{tab:ngram_samples}.

\label{app:samples}
\begin{table*}[th]
\caption{Training data examples on EMNLP2017 News and WikiText-103.}
\begin{tabular}{p{0.94\linewidth}}
  \textbf{EMNLP2017 News}  \\ \midrule\midrule
\hspace{0.5em}   My sources have suggested that so far the company sees no reason to change its tax structures , which are perfectly legal . \\
\hspace{0.5em} I ' d say this is really the first time I ' ve had it in my career : how good I feel about my game and knowing where it ' s at . \\
\hspace{0.5em} We would open our main presents after lunch ( before the Queen ' s speech ) then take the dog for a walk . \\
  \textbf{WikiText-103}  \\ \midrule \midrule
\hspace{0.5em} the actual separation of technetium @-@ N from spent nuclear fuel is a long process . \\
\hspace{0.5em} she was launched on N december N , after which fitting @-@ out work commenced . \\
\hspace{0.5em} covington was extremely intrigued by their proposal , considering eva perón to be a non @-@ commercial idea for a musical .
\label{tab:data}
\end{tabular}
\end{table*}

\begin{table*}[h]
\caption{Randomly selected \ourgan~samples on EMNLP2017 News and WikiText-103.}
\begin{tabular}{p{0.94\linewidth}}
  \textbf{EMNLP2017 News}  \\ \midrule\midrule
\hspace{0.5em} We are pleased for the trust and it was incredible , our job quickly learn the shape and get on that way . \\
\hspace{0.5em} But I obviously have him with the guys , maybe in Melbourne , the players that weren ' t quite clear there . \\
\hspace{0.5em} There is task now that the UK will make for the society to seek secure enough government budget fund reduce the economy . \\
\hspace{0.5em} Keith is also held in 2005 and Ted ' s a successful campaign spokeswoman for students and a young brothers has took an advantage of operator . \\
\hspace{0.5em} Police said how a Democratic police officer , would choose the honor of alcohol and reduce his defense and foundation . \\
\hspace{0.5em} We do not go the Blues because that I spent in ten months and so I didn ' t have a great job in a big revolution .\\
\hspace{0.5em} The 28 - year - old - son Dr Price said she would have been invited to Britain for her " friend " in a lovely family .\\
\hspace{0.5em} And as long as it is lower about , our families are coming from a friend of a family . \\ \midrule \midrule
  \textbf{WikiText-103}  \\ \midrule \midrule
\hspace{0.5em} the general manager of the fa cup final was intended for the final day as a defensive drive , rather than twenty field goals . \\
\hspace{0.5em} the faces of competitive groups and visual effects were in much of the confidence of the band at \textit{UNK} 's over close circles , and as well as changing the identical elements to the computing . \\
\hspace{0.5em} a much \textit{UNK} ground was believed to convey \textit{UNK} other words , which had been \textit{UNK} writing and that he possessed receiving given powers by his \textit{UNK} transport , rather than rendered well prior to his `` collapse of the local government . \\
\hspace{0.5em} the highest viewership from the first N @.@ N \% of the debate over the current event . \\
\hspace{0.5em} the housing of the county were built in the county behind the new south park at lake london , which , as thirty @-@ two @-@ lane work used for a new property . \\
\hspace{0.5em} near a time : bootleg was used by the brazilian navy and the german <unk> copper . \\
\hspace{0.5em} the next day , curry and defenses weren ' t , with the labour government waterfall , the powerful rock \textit{UNK} , calling him heavy @-@ action , who are shy to refuse to fight while preferred desperate oppression in alkan .
 \\
\hspace{0.5em} the british the united states launched double special education to its N \% ;
\end{tabular}
\label{tab:more_samples}
\end{table*}

\begin{table*}[h]
\caption{Randomly selected \ourgan~samples on EMNLP2017 News as training progresses.}
\begin{tabular}{p{0.94\linewidth}}
  \textbf{Beginning of training, FED=0.54}  \\ \midrule\midrule
\hspace{0.5em} because kicking firm transparency accommodation Tim earnings While contribution once forever diseases O spotlight furniture intervention guidelines false Republicans Asked defeated raid - who rapid Bryant felt ago oil refused deals today dance stocks Center reviews Storm residents emerging Duke blood draw chain Law expanding code few MPs stomach <unk> countries civilians \\
\hspace{0.5em} March labour leave theft afterwards coach 1990 importance issues American revealing players reports confirmed depression crackdown Green publication violence keeps 18th address defined photos experiencing implemented Center shots practical visa felt tweeted hurt Raiders lies artist 1993 reveal cake Amazon express party although equal touch Protection performance own rule Under golden routine \\
\midrule \midrule
  \textbf{During training, FED=0.034}  \\ \midrule \midrule
\hspace{0.5em} Cuba owners might go him because a break in a very small - defeat City drive an Commons \\
\hspace{0.5em} Germany made it by the chairman of his supporters , who are closed in Denver and 4 average - \\
\hspace{0.5em} Nine news she scored Donald Trump , appeared to present a New - \\
\hspace{0.5em} If he did , he wants a letter of the electorate that he accepted the nomination campaign for his first campaign to join passing the election . \\
\hspace{0.5em} The former complaint she said : " whatever this means certain players we cannot have the result of the current market . \\
\midrule \midrule
  \textbf{End of training, FED=0.018}  \\ \midrule \midrule
\hspace{0.5em} She ' s that result she believes that for Ms . Marco Rubio ' s candidate and that is still become smaller than ever .  \\
\hspace{0.5em} I hadn ' t been able to move on the surface -- if grow through ,' she said , given it at a time later that time .  \\
\hspace{0.5em} If Iran wins business you have to win ( Iowa ) or Hillary Clinton ' s survived nothing else since then , but also of all seeks to bring unemployment . \\
\hspace{0.5em} All the storm shows is incredible , most of the kids who are telling the girls the people we ' re not turning a new study with a challenging group . \\
\hspace{0.5em} Six months before Britain were the UK leaving the EU we will benefit from the EU - it is meeting by auto , from London , so it ' s of also fierce faith Freedom . \\
\end{tabular}
\label{tab:samples_vs_training}
\end{table*}

\begin{table*}[h]
\caption{Randomly selected MLE model samples on EMNLP2017 News and WikiText-103.}
\begin{tabular}{p{0.94\linewidth}}
  \textbf{EMNLP2017 News}  \\ \midrule\midrule
\hspace{0.5em} It came out the five days of the developing player waiting to begin the final major European championship of state - owned teams in 2015 and 2015 . \\
\hspace{0.5em} " I look from my size , you know in the most part , I ' ve been fighting every day , " she says . \\
\hspace{0.5em} When you are around mid - 2006 , you play one and train with you earlier this year and the manager would make the opposition . \\
\hspace{0.5em} She said : ' I ' d like food to be now , where my baby and children deserve to be someone ' s kids . \\
\hspace{0.5em} He ' d been very good at that , but it ' s fun , the camera have been incredibly tight - with that we can be on the ball at the beginning of his debut . \\ \midrule \midrule
  \textbf{WikiText-103}  \\ \midrule \midrule
\hspace{0.5em} in an interview with journalist \textit{UNK} \textit{UNK} during his death , a new specimen was brought in the \textit{UNK} museum of modern art . \\
\hspace{0.5em} after the sets of \textit{UNK} wear \textit{UNK} and \textit{UNK} ' \textit{UNK} ' \textit{UNK} to tell him , \textit{UNK} \textit{UNK} they play \textit{UNK} \textit{UNK} with \textit{UNK} around a \textit{UNK} . \\
\hspace{0.5em} after he urged players to fight for what he saw as a fantastic match , the bank sustained a fractured arm and limited injury in the regular season . \\
\hspace{0.5em} the album peaked at number eight on rolling stones ' s N . \\
\hspace{0.5em} in the \textit{UNK} sitting on the starboard N @-@ inch , a \textit{UNK} woman looks ( \textit{UNK} \textit{UNK} ) with an eagle during the day of all singing due to her the doors being edged far through where she \textit{UNK} , which included \textit{UNK} , \textit{UNK} \textit{UNK} , \textit{UNK} \textit{UNK} , \hspace{0.5em} and \textit{UNK} 's motifs on the bridge .
\end{tabular}
\label{tab:mle_samples}
\end{table*}

\begin{table*}[h]
\caption{Randomly selected samples from an $5$-gram model with Kneser-Ney smoothing.}
\begin{tabular}{p{0.94\linewidth}}
  \textbf{EMNLP2017 News}  \\ \midrule\midrule
\hspace{0.5em} It ' s like a ' test site will boost powerful published on the question , 60 years on the fact that at moment . \\
\hspace{0.5em} The bridge opens fire Dallas - and they ' ll be best remembered as scheduled by accident and emergency units . \\
\hspace{0.5em} The study focused on everything Donald Trump was " somebody to cope with a social events that was not wearing the result of a 1 , 2017 , will be in . \\
\hspace{0.5em} It ' s going to finish me off , when a recent poll , more than the actual match to thank the British way of the seven years . \\
\hspace{0.5em} We can be sure that has been struck off by the company , is to be completed by Smith had taken a week later , you just like , what ' s going on in everyday reflects a material drone hundreds of comments .
\end{tabular}
\label{tab:ngram_samples}
\end{table*}

\end{document}